\pdfoutput=1

\documentclass[11pt]{article}

\usepackage[]{acl}

\usepackage{times}
\usepackage{latexsym}

\usepackage[T1]{fontenc}

\usepackage[utf8]{inputenc}

\usepackage{microtype}

\usepackage{inconsolata}

\usepackage{graphicx}

\usepackage[utf8]{inputenc}
\usepackage[T1]{fontenc}
\usepackage{algorithm}
\usepackage{algorithmic}
\usepackage{graphicx}
\usepackage{array}
\usepackage{booktabs}
\usepackage{multirow}
\usepackage{makecell}
\usepackage{amsmath}
\usepackage{amsfonts}
\usepackage{enumitem}
\usepackage{url} 
\usepackage{xspace}
\usepackage{xcolor}
\usepackage{hyperref}
\usepackage{epigraph}
\usepackage{tabularx}
\usepackage{colortbl}

\definecolor{langblue}{rgb}{0, 0.4, 0.8}
\definecolor{langred}{rgb}{0.81, 0.09, 0.13}
\definecolor{langgreen}{rgb}{0.0, 0.6, 0.3}
\hypersetup{
}

\makeatletter
\AtBeginDocument{%
  \let\oldref\ref
  \renewcommand{\ref}[1]{%
    \hyperref[{#1}]{\underline{\oldref{#1}}}%
  }%
}
\makeatother

\usepackage{titletoc}
\newcommand\DoToC{%
  \startcontents
  \printcontents{}{1}{\textbf{Contents of Appendix}\vskip3pt\hrule\vskip5pt}
  \vskip3pt\hrule\vskip5pt
}

\newcommand{\method}{TablePilot\xspace}
\newcommand{\dataset}{DART\xspace}
\newcommand{\alignmethod}{Rec-Align\xspace}

\title{\method: Recommending Human-Preferred Tabular Data Analysis \\ with Large Language Models}

\author{
Deyin Yi\textsuperscript{\rm 1}\thanks{\ \ Work during internship at Microsoft.},
Yihao Liu\textsuperscript{\rm 2}\footnotemark[1],
Lang Cao\textsuperscript{\rm 3}\footnotemark[1],
\\
\textbf{
Mengyu Zhou\textsuperscript{\rm 4}\thanks{\ \ Corresponding author (mezho@microsoft.com).},
Haoyu Dong\textsuperscript{\rm 4},
Shi Han\textsuperscript{\rm 4},
Dongmei Zhang\textsuperscript{\rm 4}
}
\\
\textsuperscript{\rm 1}Shanghai University of Finance and Economics \textsuperscript{\rm 2}Peking University \\
\textsuperscript{\rm 3}University of Illinois Urbana-Champaign \textsuperscript{\rm 4}Microsoft Research \\
}

\begin{document}
\maketitle
\begin{abstract}
Tabular data analysis is crucial in many scenarios, yet efficiently identifying the most relevant data analysis queries and results for a new table remains a significant challenge. The complexity of tabular data, diverse analytical operations, and the demand for high-quality analysis make the process tedious. To address these challenges, we aim to recommend query–code–result triplets tailored for new tables in tabular data analysis workflows. In this paper, we present \method, a pioneering tabular data analysis framework leveraging large language models to autonomously generate comprehensive and superior analytical results without relying on user profiles or prior interactions. The framework incorporates key designs in analysis preparation and analysis optimization to enhance accuracy. Additionally, we propose \alignmethod, a novel method to further improve recommendation quality and better align with human preferences. Experiments on \dataset, a dataset specifically designed for comprehensive tabular data analysis recommendation, demonstrate the effectiveness of our framework. Based on GPT-4o, the tuned \method achieves 77.0\% top-5 recommendation recall. Human evaluations further highlight its effectiveness in optimizing tabular data analysis workflows.
\end{abstract}

\section{Introduction}
\begin{figure}[t]
    \centering
    \includegraphics[width=1\linewidth]{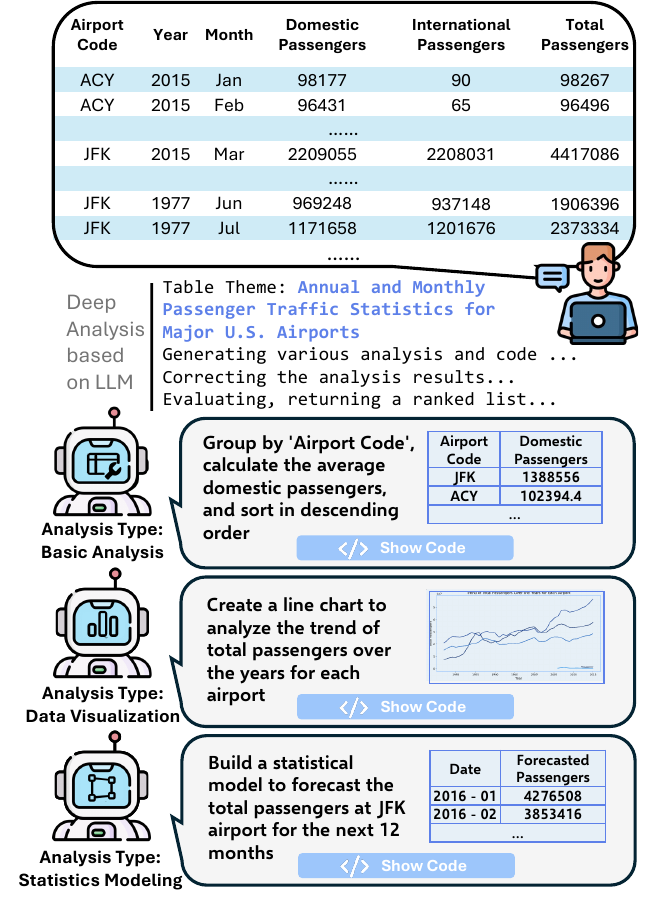}
    \caption{Overview of \method. Through deep analysis based on LLM, \method generates three types of analysis: basic analysis, data visualization, and statistical modeling, each presented as a <query, code, result> triplet.}
    \label{fig:overview}
\end{figure}

Tabular data is widely used in various data analysis scenarios \cite{ghasemi2016process,li2021gfte}. However, its complexity and density \cite{cao2025tablemasterrecipeadvancetable,tian2024spreadsheetllm} can make it challenging, even for professional analysts, to determine the most appropriate analysis operations for a new table. Conducting tabular data analysis is often tedious, and the analysis operations may include errors that lead to suboptimal outcomes. Therefore, automatically recommending high-quality analysis queries and results becomes essential in the data analysis workflow, particularly in zero-turn scenarios where no user profile or historical records are available.

In the task of tabular data analysis recommendation, given only a table as input, we aim to recommend query–code–result triplets to users. A query specifies the type and objective of the analysis task, while the code executes the corresponding operation on the table, serving as an intermediate step in the analysis. The result presents the execution output, which also constitutes the analysis findings.

Previous works on tabular data analysis recommendation \cite{zhou2021table2charts,zhou2020table2analysis} primarily rely on traditional machine learning methods. However, these approaches suffer from poor generalization ability, suboptimal performance, and a strong dependence on specific datasets. Recently, large language models (LLMs) \cite{openai2024gpt4technicalreport, touvron2023llamaopenefficientfoundation} have achieved significant progress in natural language processing. With their strong data processing, language understanding, and generation capabilities, LLMs offer new opportunities to deliver more effective tabular analysis recommendations to users.

In practical data analysis scenarios, these triplets are expected to be (a) \textbf{accurate}, (b) \textbf{diverse}, and (c) \textbf{human-preferred}. Human-preferred is a general term that refers to results aligning with human values and preferences, meaning they should be meaningful, insightful, interpretable, and so on. Employing LLMs to recommend tabular data analyses while meeting these requirements presents several key challenges.

\textit{Challenges:} (a) Tabular data is often large and data-intensive, making it difficult for LLMs to process effectively. Long-context windows can trigger hallucinations \cite{huang2024hallucination}, leading to inaccurate results. (b) Existing approaches primarily construct workflows for a single type of operation, executing predefined data analysis queries to obtain results \cite{fang2024large,zhang2025survey}. However, these methods lack diversity and fail to provide a comprehensive analysis. Additionally, designing a system that effectively combines and delivers the final analysis results to users is also challenging. (c) Selecting and presenting analysis results in a way that aligns with human cognitive patterns is crucial \cite{song2024preference, dai2023safe, yu2024rlhf}. A well-designed system should recommend a diverse set of data analysis operations that match users’ analytical preferences, ensuring that the generated insights are interpretable, actionable, and meaningful.

\textit{Solution:} To address these challenges, we propose \textbf{\method}, a framework designed to tackle the zero-turn recommendation task for tabular data analysis, as illustrated in Figure~\ref{fig:overview}. 

To enhance the \textbf{accuracy} of analysis results, we adopt sampling techniques \cite{sui2024table, ye2023large, ji2024tree}, employing a table sampler to refine model inputs and introducing a table explanation component that incorporates world knowledge learned during the pretraining phase of LLMs. This stage of analysis preparation facilitates the generation of more contextually appropriate queries and results. At the optimization level, we utilize post-refinement techniques \cite{chen2022large, he2024cocost} to adjust outputs. However, instead of focusing solely on code refinement, we identify multiple aspects of query and result optimization.

To improve the \textbf{diversity} of our analysis, we implement a modularized approach to support various workflow operations. This modular design provides two key benefits. First, it ensures comprehensive coverage by enabling the workflow to handle a diverse range of data analysis tasks, making it more adaptable to various requirements. Second, it enhances performance by allowing each module to be trained independently for better efficiency, with improvements across modules contributing to overall effectiveness.

To ensure our analysis aligns with \textbf{human preferences}, we introduce \textbf{\alignmethod}, a method specifically designed to further enhance the quality of analysis by directly incorporating human preferences. We train a ranking model to optimize the final set of recommended operations, ensuring they align with human analytical tendencies and produce superior results.

We contribute a dataset \textbf{\dataset} to support and validate our framework. Experimental results demonstrate that the tuned \method achieves nearly 100\% execution rates, while the analysis modules show an overall recall improvement of 11.25\% with GPT-4o in the dataset. \alignmethod further enhances alignment with human preferences, leading to gains of 6.8\% in Recall@3 and 6.0\% in Recall@5. Additionally, human evaluations confirm that the \method framework provides more practical and insightful data analysis recommendations compared to baseline models. Extensive experiments validate the effectiveness of \method and our training approach.

In summary, our main technical contributions are as follows:
\begin{itemize}[leftmargin=*, itemsep=0pt, labelsep=5pt, topsep=0pt]
    \item We propose \method, a framework designed to tackle the zero-turn recommendation task for tabular data analysis. The core analysis follows a modular design to ensure diversity in the final analysis. We also contribute \dataset, a dataset to support and validate our framework.
    \item We introduce two additional steps to enhance the accuracy of analysis results, applied before and after the core analysis. These steps incorporate sampling, explanation, and multi-faceted refinement.
    \item We develop \alignmethod, a method designed to align recommendations with human analytical preferences, thereby further enhancing the quality and practical utility of the recommended results.
\end{itemize}

\section{Related Work}
Current tabular data analysis recommendation tasks can be categarized into three main types:
\\

\noindent\textbf{Basic Data Analysis in Tables.}
Basic analysis refers to simple, initial processing of a table. It involves generating tabular outputs or single-cell text entries to highlight key information or insights based on a user query. This is usually done by manipulating and aggregating tabular data. Table understanding tasks \cite{pasupat2015compositionalsemanticparsingsemistructured,2019TabFactA} are the most basic form of this analysis. Given a query, these tasks either provide an answer or extract a sub-table \cite{wang2024chainoftableevolvingtablesreasoning, ye2023largelanguagemodelsversatile} that contains important information. TableMaster \cite{cao2025tablemasterrecipeadvancetable} offers a general recipe for table understanding and basic analysis based on user queries. Text2SQL \cite{pourreza2024din, gao2023text, lee2024mcs,zhao2024nl2formula} is another approach that extracts relevant parts of a table by converting user queries into SQL-based outputs. However, these methods only return results based on a given query and do not generate natural language queries automatically. Auto-Formula \cite{chen2024auto} predicts and suggests formula syntax for spreadsheet-based analysis. Table2Analysis \cite{zhou2020table2analysis} and MetaInsight \cite{ma2021metainsight} automatically recommends common analysis without requiring user input.
\\

\noindent\textbf{Tabular Data Visualization.}
Visualizing data helps users quickly understand complex patterns and relationships. Table2Charts \cite{zhou2021table2charts} applies sequence token sampling and reinforcement learning to recommend different chart types. Furmanova et al. \cite{Furmanova_2019} developed a tool for automatically combining overview and details in tabular data visualizations. AdaVis \cite{zhang2023adavisadaptiveexplainablevisualization} uses knowledge graphs to adaptively recommend one or multiple suitable visualizations for a dataset. LLMs have further improved data visualization. Chart2VIS \cite{maddigan2023chat2vis} leverages LLMs for natural language-to-visualization tasks by generating Python code for chart creation. ChartLlama \cite{han2023chartllama}, a multi-modal LLM, shows strong chart generation capabilities but does not recommend charts based on existing data.
\\

\noindent\textbf{Statistical Modeling of Tabular Data}
Statistical modeling in tabular data focuses on building models to recognize patterns and relationships. RIM \cite{qin2021retrieval} enhances tabular data prediction with a retrieval module. GReaT \cite{borisov2022language} uses a decoder-only transformer to model data distributions and generate realistic synthetic data. GTL \cite{wen2024supervised} integrates LLMs with deep learning techniques for regression and classification tasks. TabDDPM \cite{kotelnikov2024tabddpmmodellingtabulardata} is a diffusion model that can handle any tabular dataset and support various feature types.
\\

Despite these advancements, most existing methods are task-specific and do not support multiple types of analysis within a single framework. This limitation prevents users from obtaining a comprehensive view of their data. Currently, no unified system seamlessly integrates table analysis, visualization, and statistical modeling. A complete all-in-one framework would allow users to explore data more effectively from different perspectives. Moreover, existing methods primarily emphasize the accuracy of analysis results while neglecting the importance of aligning with human analytical preferences.




\section{Methodology}
\label{sec:method}
\begin{figure*}
    \centering
    \includegraphics[width=1\linewidth]{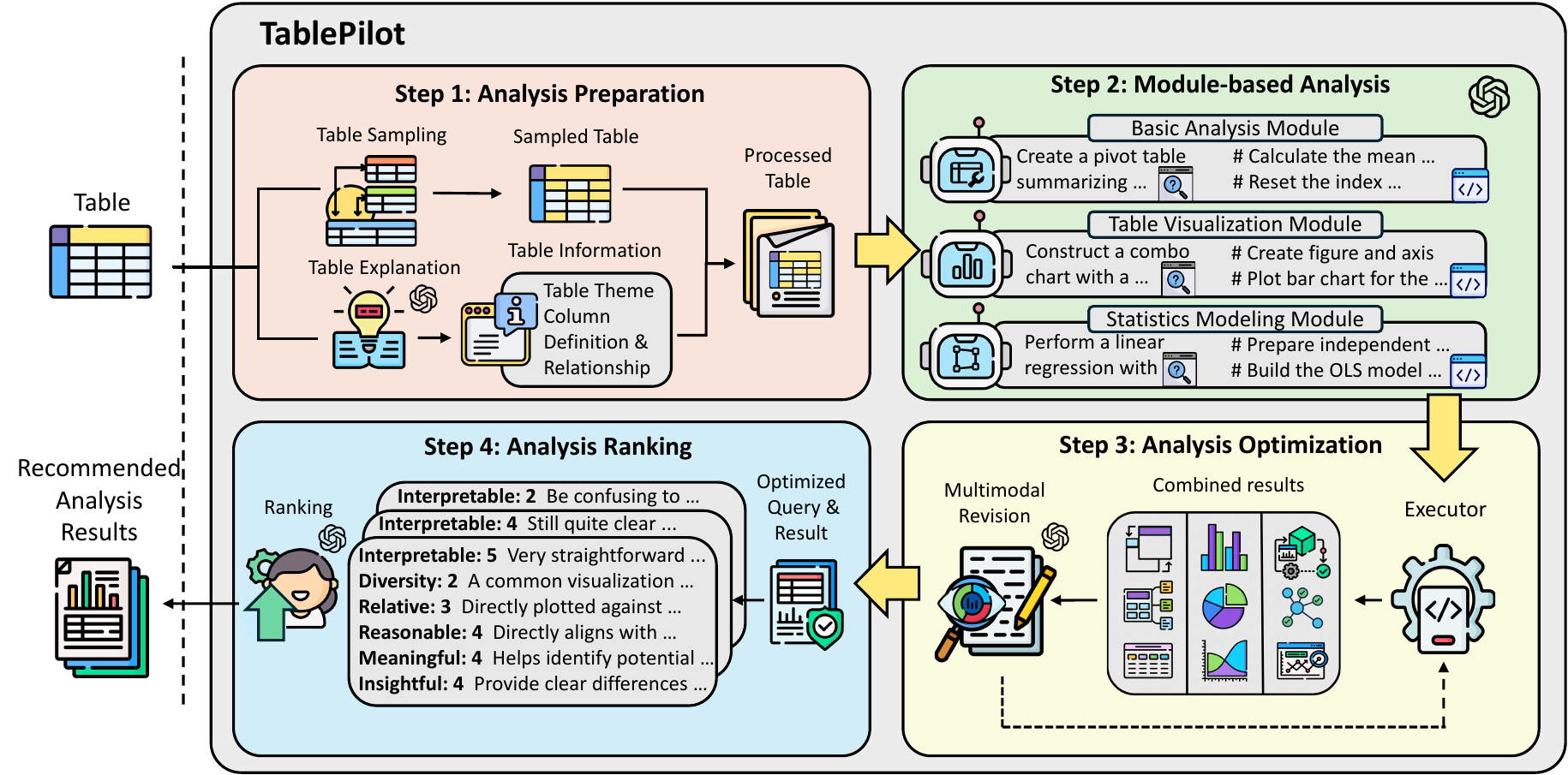}
    \caption{The \method framework. Step 1: Sample the input table and generate corresponding explanations for its structure and content. Step 2: Generate query and code for modules involving basic analysis, table visualization, and statistics modeling. Step 3: Optimize the quality of <query, code, result> triplets. Step 4: Score and rank the optimized results based on multiple criteria to recommend the top-K analysis. \method Case Study and Analysis Report can be seen at Appendix~\ref{ap:case study} and Appendix~\ref{ap:report}.}
    \label{fig:framework}
\end{figure*}

\subsection{Task Formulation}
\textbf{Tabular Data Analysis Recommendation.}
In the task of tabular data analysis recommendation, the objective is to generate a series of recommended data analysis queries $q$, corresponding code $c$, and execution results $r$ for a given table $\mathbb{T}$ under a zero-turn setting (i.e., with no user profile or historical context). The table $\mathbb{T}_{a\times b}$ contains $a$ rows and $b$ columns, where $C_{i,j}$ denotes the cell in the $i$-th row and $j$-th column. For each table $\mathbb{T}$, $n$ analysis results $A$ is recommended in triplets:
\begin{equation}
A \;=\; \bigl\{\,\bigl(q_{i},\,c_{i},\,r_{i}\bigr)\bigr\}_{i=1}^{n},
\end{equation}
where each triplet $a = (q,\,c,\,r)$ represents a single recommended analysis result.

\subsection{Framework}
To generate recommendation results from a given table, we propose \method, a four-step analysis framework, as illustrated in Figure~\ref{fig:framework}. The framework consists of Analysis Preparation, Module-based Analysis, Analysis Optimization, and Analysis Ranking. A new table $T$ is provided as input to generate the recommended results $A$.
\begin{equation}
\text{\method}(\mathbb{T}) = A.
\end{equation}

\noindent\textbf{Step 1: Analysis Preparation.} The objective of this step is to transform raw tabular data into a more focused form that facilitates efficient analysis. This step involves two key tasks: sampling a subset of the table and generating a table explanation.

Raw tables often contain large amounts of data, much of which may not be relevant for a specific analysis task. Sampling extracts a representative subset of the table, capturing essential patterns while reducing computational load and focusing the analysis on key data points. This process involves selecting a subset of rows from the original table:
\begin{equation}
\text{Sampling}(\mathbb{T}_{a\times b}) = \mathbb{T}'_{a'\times b'},
\end{equation}
where $\mathbb{T'}$ represents the sampled table, $a'$ denotes the number of selected rows, and $b'$ denotes the number of selected columns.

Additionally, generating a table explanation is crucial for structuring the data, making column relationships and the table’s overall theme clearer and more interpretable. This explanation includes metadata such as the table’s theme, column descriptions, and relationships between different columns, all of which guide subsequent analysis. The explanation is denoted as $E$:

\begin{equation}
\text{Explanation}(\mathbb{T}) = E.
\end{equation}

\noindent\textbf{Step 2: Module-based Analysis.}
In this step, we perform a module-based analysis on the sampled table $\mathbb{T}’$ and its corresponding table explanation $E$. The goal is to generate analysis results by applying specialized modules to different aspects of the data. These modules focus on basic analysis ($ba$), data visualization ($dv$), and statistical modeling ($sm$). Each module takes $\mathbb{T}’$ and $E$ as inputs to generate meaningful query-code pairs $(q, c)$:
\begin{equation}
\mathcal{M}_k(\mathbb{T}’, E) = (q_k, c_k),
\end{equation}
where $k \in \{ba, dv, sm\}$ represents the three different analysis task.

The Basic Analysis module ($\mathcal{M}_{ba}$) applies fundamental yet powerful techniques to explore the data, performing operations such as filtering, grouping, sorting, and aggregation. The Data Visualization module ($\mathcal{M}_{dv}$) generates visual representations of the data to reveal patterns, trends, and relationships. The Statistical Modeling module ($\mathcal{M}_{sm}$) applies advanced statistical techniques to analyze the data and uncover deeper insights. It may involve regression analysis, hypothesis testing, or predictive modeling, depending on the analysis objectives.

\noindent\textbf{Step 3: Analysis Optimization}
In this step, we first execute the code to obtain results $r$ for each code $c_k$:
\begin{equation}
\text{Execution}(\mathbb{T}, c_k) = r = 
\begin{cases}
T, & \text{if } k = ba \\
V, & \text{if } k = dv \\
M, & \text{if } k = sm
\end{cases}
\end{equation}
where $T$ represents the sub-table after data manipulation in basic analysis, $V$ denotes the result of data visualization, and $M$ corresponds to the output of statistical modeling. The result of data visualization, $r = V$, is also an image $r = I$, which will be used as input for the vision module of LLMs at a later stage. We then combine the query $q$, code $c$, and result $r$ into an analysis triplet $a = (q, c, r)$. The results  $r = \text{Error}$ indicate an error in the code execution.

Next, we refine the analysis triplet $a$  based on the results from table sampling $\mathbb{T}$ and explanation $E$. The optimization process utilizes LLMs to improve the alignment of queries and code with the data and analysis intent, ensuring more accurate and meaningful results. There are two different strategies for LLMs to optimize triplets, depending on whether the result contains an error. After refinement, the optimized code is executed to generate the final enhanced execution results, yielding an optimized triplet $a’ = (q’, c’, r’)$:

\begin{equation}
a' =
\begin{cases}
\text{Optimize}_A(q, c, r \mid \mathbb{T}, E), & \text{if } r \neq \text{Error} \\
\text{Optimize}_B(q, c, r \mid \mathbb{T}, E), & \text{if } r = \text{Error}
\end{cases}
\end{equation}

\noindent\textbf{Step 4: Analysis Ranking}
In the final step, the objective is to evaluate and rank all the $(q, c, r)$ triplets $A = {a_i}_{i=1}^{n}$ that were generated and optimized in the previous step. To achieve this, we design a ranking module that scores each triplet based on multiple dimensions, such as relevance, diversity, and other key factors (criteria detailed in Appendix~\ref{ap:criteria}). These scores are then aggregated to compute an overall score $s$. Using these scores, the triplets are ranked in descending order, allowing us to select the top $k$ results:
\begin{equation}
A'_{k} = \text{Top}_k \Big( \text{Rank} \Big( \big\{ (q', c', r')_i \big\}_{i=1}^{n} \Big) \Big)
\end{equation}

After scoring, ranking, and selecting the top-$k$ results $A’_{k}$, the final triplets are recommended to users.
       
\subsection{Training}
The training process in \method is designed to enhance the model’s ability to generate high-quality analysis results, with a focus on accurate query-code generation and human-preferred ranking of analysis triplets $a = (q,c,r)$. We primarily employ Supervised Fine-Tuning (SFT) and Direct Preference Optimization (DPO) (introduced in Appendix~\ref{ap:dpo}), both widely used techniques for tuning LLMs. SFT is used to ensure that each module follows our instructions for performing tasks. Additionally, we introduce \textbf{\alignmethod}, implemented via DPO, to enhance our ranking module, further refining recommendation quality and ensuring that the selected results align more closely with human preferences.

Our training strategy consists of the following key components:
\begin{itemize}[leftmargin=*, itemsep=0pt, labelsep=5pt, topsep=1pt]
    \item \textbf{Analysis SFT} trains the LLMs in three analysis module ($\mathcal{M}_{ba}$, $\mathcal{M}_{dv}$, $\mathcal{M}_{sm}$) to improve their ability to follow instructions, generating relevant queries and accurate code. This enhances the accuracy of the analysis.
    \item \textbf{Rank SFT} trains the LLMs in the ranking module \textit{Rank} to better follow instructions in evaluating each analysis triplet based on comprehensive criteria and assigning appropriate scores. This ensures that the ranking model adheres to our guidelines when ranking triplets..
    \item \textbf{Rank DPO} implements \alignmethod through DPO to refine the evaluation of analysis triplets in \textit{Rank}, ensuring that evaluation and scoring are more closely aligned with human analytical preferences. This further enhances the quality of the recommended analysis.
\end{itemize}

\section{Experiments}
\begin{table*}[t!]
\centering
\resizebox{\textwidth}{!}{%
\begin{tabular}{l|ccccccccc|ccc}
\toprule
\multirow{2}{*}{\textbf{Method}}
& \multicolumn{3}{c}{\textbf{Basic Analysis}}
& \multicolumn{3}{c}{\textbf{Data Visualization}}
& \multicolumn{3}{c}{\textbf{Statistics Modeling}}
& \multicolumn{3}{|c}{\textbf{Overall}} \\
\cmidrule(lr){2-4}\cmidrule(lr){5-7}\cmidrule(lr){8-10}\cmidrule(lr){11-13}
& \textbf{R@3} & \textbf{R@5} & \textbf{R@N}
& \textbf{R@3} & \textbf{R@5} & \textbf{R@N}
& \textbf{R@3} & \textbf{R@5} & \textbf{R@N}
& \textbf{R@3} & \textbf{R@5} & \textbf{R@N} \\
\midrule
\makecell[l]{\textbf{GPT-4o}} &  &  & &  &  & &  &  & &  &  &  \\
\makecell[l]{\quad Baseline}
 & 13.00 & 20.11 & 42.00
 & 17.57 & 26.30 & 53.40
 & 15.08 & 27.08 & 56.67
 & 38.11 & 52.11 & 80.00 \\

\makecell[l]{\quad Vanilla}
 & 14.05 & 21.07 & 50.67
 & 35.84 & 48.81 & 69.37
 & 15.48 & 38.91 & 59.58
 & 53.51 & 70.90 & 87.67 \\

\makecell[l]{\quad Analysis\ SFT + Rank Vanilla}
 & 15.67 & 22.33 & 55.33
 & 43.88 & 53.06 & 70.41
 & 20.00 & 30.42 & 61.25
 & 59.00 & 72.67 & 89.00 \\

\makecell[l]{\quad Analysis\ SFT + Rank SFT}
 & 15.67 & \underline{28.00} & 55.33
 & 41.84 & 53.06 & 70.41
 & \underline{21.25} & 38.33 & 61.25
 & 58.00 & 74.33 & 89.00 \\

\makecell[l]{\quad Analysis\ SFT + Rank SFT-V}
 & 15.33 & 25.67 & 55.33
 & \textbf{44.22} & \underline{54.42} & 70.41
 & 16.25 & \underline{45.83} & 61.25
 & 61.00 & 75.00 & 89.00 \\

\makecell[l]{\quad Analysis\ SFT + Rank SFT \& DPO}
 & \textbf{19.33} & \textbf{30.00} & 55.33
 & 42.86 & 52.72 & 70.41
 & 20.42 & 42.08 & 61.25
 & \underline{61.33} & \underline{76.00} & 89.00 \\

\rowcolor{gray!20}
\makecell[l]{\quad Analysis\ SFT + Rank SFT-V \& DPO}
 & \underline{17.67} & 26.00 & 55.33
 & \underline{43.88} & \textbf{54.78} & 70.41
 & \textbf{22.92} & \textbf{47.08} & 61.25
 & \textbf{63.00} & \textbf{77.00} & 89.00 \\
\midrule

\makecell[l]{\textbf{GPT-4o-mini}} &  &  & &  &  & &  &  & &  &  &  \\
\makecell[l]{\quad Baseline}
 & 15.99 & 24.94 & 35.33
 & 27.33 & 39.33 & 44.22
 & 3.61 & 6.67 & 35.33
 & 29.33 & 42.44 & 62.67 \\

\makecell[l]{\quad Vanilla}
 & 8.67 & 10.67 & 38.33
 & 40.48 & 50.34 & 56.12
 & 5.54 & 10.83 & 38.33
 & 45.33 & 56.67 & 78.33 \\

\makecell[l]{\quad Analysis\ SFT + Rank Vanilla}
 & 13.00 & \textbf{57.14} & 46.67
 & \underline{44.22} & 25.33 & 64.29
 & 1.67 & 10.42 & 59.58
 & 52.00 & 68.67 & 85.00 \\

\makecell[l]{\quad Analysis\ SFT + Rank SFT}
 & \textbf{24.91} & \underline{34.33} & 46.67
 & 34.15 & 45.24 & 64.29
 & 12.02 & 32.08 & 59.58
 & 56.66 & 71.67 & 85.00 \\

\makecell[l]{\quad Analysis\ SFT + Rank SFT-V}
 & 16.00 & 24.33 & 46.67
 & \textbf{46.60} & \textbf{54.08} & 64.29
 & \underline{22.50} & \underline{43.33} & 59.58
 & \underline{61.00} & \underline{75.00} & 85.00 \\

\makecell[l]{\quad Analysis\ SFT + Rank SFT \& DPO}
 & \underline{21.33} & 32.67 & 46.67
 & 42.86 & 50.34 & 64.29
 & 16.25 & 27.05 & 59.58
 & 60.33 & 73.67 & 85.00 \\

\rowcolor{gray!20}
\makecell[l]{\quad Analysis\ SFT + Rank SFT-V \& DPO}
 & 21.00 & 29.00 & 46.67
 & 40.14 & \underline{51.02} & 64.29
 & \textbf{22.92} & \textbf{49.17} & 58.58
 & \textbf{62.33} & \textbf{76.67} & 85.00 \\
\midrule

\makecell[l]{\textbf{Phi-3.5-vision}} &  &  & &  &  & &  &  & &  &  &  \\
\makecell[l]{\quad Baseline}
 & 3.00 & 4.00 & 5.00
 & 1.36 & 3.40 & 4.08
 & 0.00 & 0.00 & 0.42
 & 4.33 & 7.00 & 8.67 \\

\makecell[l]{\quad Vanilla}
 & 1.43 & 1.79 & 13.33
 & 1.83 & 1.83 & 3.74
 & 3.12 & 3.12 & 7.92
 & 5.73 & 6.09 & 21.67 \\

\makecell[l]{\quad Analysis\ SFT + Rank Vanilla}
 & 3.77 & 3.77 & 24.00
 & \textbf{3.83} & \textbf{4.53} & 9.52
 & 18.45 & 19.31 & 32.50
 & 20.89 & 21.58 & 47.67 \\

\makecell[l]{\quad Analysis\ SFT + Rank SFT}
 & \textbf{6.85} & 14.04 & 24.00
 & \underline{2.79} & \underline{4.18} & 9.52
 & 15.88 & \underline{22.75} & 32.50
 & 20.89 & 32.19 & 47.67 \\

\makecell[l]{\quad Analysis\ SFT + Rank SFT-V}
 & 5.14 & 13.01 & 24.00
 & 1.74 & 3.14 & 9.52
 & \textbf{19.31} & 21.89 & 32.50
 & 21.23 & 30.14 & 47.67 \\

\makecell[l]{\quad Analysis\ SFT + Rank SFT \& DPO}
 & \textbf{8.90} & \textbf{15.07} & 24.00
 & 1.74 & 3.83 & 9.52
 & 18.88 & 23.61 & 32.50
 & \textbf{23.97} & \textbf{32.88} & 47.67 \\

\rowcolor{gray!20}
\makecell[l]{\quad Analysis\ SFT + Rank SFT-V \& DPO}
 & \underline{7.53} & \underline{14.38} & 24.00
 & 1.74 & 2.09 & 9.52
 & \textbf{19.31} & \textbf{25.32} & 32.50
 & \underline{23.63} & \underline{32.19} & 47.67 \\
\bottomrule
\end{tabular}
}
\caption{Recall across multiple models and experimental settings (all values in \%). Experimental results demonstrate the effectiveness of \method, with \textit{Analysis SFT + Rank SFT-V \& DPO} generally achieving the best performance.}
\label{tab:main_recall}
\end{table*}

We conduct extensive experiments and provide a detailed analysis to demonstrate the effectiveness of \method. To support, validate the framework, and evaluate its performance, we carefully curate a dataset, \textbf{\dataset} (representing \textbf{D}ata \textbf{A}nalysis \textbf{R}ecommendation for \textbf{T}ables). Details on its construction and further information of our dataset can be found in Appendix~\ref{ap:dataset}.

\subsection{Experiment Settings}
In our experiments, we evaluate the performance of \method on three typical analysis tasks: Basic Analysis, Data Visualization, and Statistics Modeling. Additionally, we consider them collectively without distinction for the overall evaluation. We aim to evaluate the result of query–code–result triplets for a given table. To assess the quality of code generation, we use the execution rate as a metric. For the quality of the final results in recommendation, we evaluate using Recall@K. Detailed evaluation metrics can be found in Appendix~\ref{ap:metric}.

We selected three state-of-the-art vision-language models of varying sizes and availability, including both private and open-source options, as the foundation models: \textit{Phi-3.5-Vision}, \textit{GPT-4o}, and \textit{GPT-4o-mini}. These models were chosen for their strong vision-language interaction capabilities, making them well-suited for multi-modal refinement. \textit{GPT-4o} and \textit{GPT-4o-mini} are both recognized for their strong performance in vision-language tasks. While \textit{GPT-4o} provides high-quality outputs, \textit{GPT-4o-mini} serves as a more efficient alternative with minimal trade-offs in accuracy, enabling a balanced comparison for evaluating our framework. \textit{Phi-3.5-Vision} is an open-source model, ensuring easy reproducibility and facilitating future research.

We conduct multiple comparative experiments to comprehensively evaluate performance. Specifically, the \textit{baseline} experiments do not incorporate any components of our proposed framework; instead, they use a single prompt to generate queries and code for all three task categories simultaneously, with recall computed using random ranking. In contrast, the \textit{vanilla} experiments utilize \method without any additional model tuning. The remaining experiments explore various components within \method, applying different tuning methods such as SFT and DPO. The meanings of \textit{Analysis SFT}, \textit{Rank SFT}, and \textit{Rank DPO} are detailed in Section~\ref{sec:method}. \textit{Rank Vanilla} refers to random ranking. Additionally, the \textit{-V} notation indicates whether vision input is used during the training stage.

We then compare the outcomes across these experimental settings to comprehensively assess the contributions of each tuning strategy to the overall performance improvements. More detailed settings can be found in Appendix~\ref{ap:settings}. The prompts are provided in Appendix~\ref{ap:prompt}.

\begin{table}[t!]
\centering
\resizebox{0.48\textwidth}{!}{%
\begin{tabular}{l|ccc}
\toprule
\multirow{2}{*}{\textbf{Method}}
& \multicolumn{3}{c}{\textbf{ExecRate}} \\
\cmidrule(lr){2-4}
& \textbf{\small Basic Analysis} & \textbf{\small Data Visualization} & \textbf{\small Statistics Modeling} \\
\midrule
\makecell[l]{\textbf{GPT-4o}} &  &  &  \\
\makecell[l]{\quad Baseline}
 & 96.07 & 95.00 & 95.00 \\
\makecell[l]{\quad Vanilla}
 & 99.67 & 99.67 & \textbf{99.44} \\
\rowcolor{gray!20}
\makecell[l]{\quad Analysis\ SFT}
 & \textbf{100.00} & \textbf{99.93} & 99.33 \\
\midrule

\makecell[l]{\textbf{GPT-4o-mini}} &  &  &  \\
\makecell[l]{\quad Baseline}
 & 91.37 & 88.75 & 56.11 \\
\makecell[l]{\quad Vanilla}
 & 96.32 & 97.80 & 92.76 \\
\rowcolor{gray!20}
\makecell[l]{\quad Analysis\ SFT}
 & \textbf{99.40} & \textbf{99.66} & \textbf{98.73} \\
\midrule

\makecell[l]{\textbf{Phi-3.5-vision}} &  &  &  \\
\makecell[l]{\quad Baseline}
 & 44.17 & 26.65 & 10.83 \\
\makecell[l]{\quad Vanilla}
 & 77.03 & 57.55 & 65.78 \\
\rowcolor{gray!20}
\makecell[l]{\quad Analysis\ SFT}
 & \textbf{87.80} & \textbf{99.28} & \textbf{85.11} \\
\bottomrule
\end{tabular}
}
\caption{Execution rate across multiple models and experimental settings (all values in \%)}
\label{tab:main_execrate}
\end{table}

\subsection{Main Results}
\noindent\textbf{\method Performance.}
As illustrated in Figure~\ref{tab:main_recall} and Figure~\ref{tab:main_execrate}, \method delivers substantial performance improvements across various models without the need for fine-tuning LLMs. Notably, \textit{GPT-4o} benefits from \method, with enhancements observed across all key metrics—execution rate, recall at different thresholds (recall@3, recall@5, and recall@N) and overall task performance. The model achieves consistent gains across all tasks, underscoring the effectiveness in elevating LLM performance without requiring manual adjustments or additional tuning.

Notably, we observe some performance drops in certain analysis among basic analysis, data visualization, and statistics modeling. This is due to a \textbf{diverse analysis trade-off effect}, where an excessively high recall in one task may lead to a decline in recall for others. Therefore, overall recall serves as a more reliable measure of the method’s overall performance.

A similar improvement pattern is observed in smaller models. While their baseline performance is lower than that of \textit{GPT-4o}, they still achieve significant gains. Specifically, overall recall@N for \textit{GPT-4o-mini} and \textit{Phi-3.5-vision} increases by 16\% and 13\%, respectively, with execution rates also improving—some tasks see up to a 50\% boost, particularly in modeling tasks. These results confirm that, regardless of model size, \method consistently enhances recall and execution reliability.

\noindent\textbf{\method Performance after Tuning.}
Supervised Fine-Tuning significantly improves both analysis and ranking tasks. Vision-enabled SFT further enhances ranking performance, especially when combined with DPO applied to vision components. While \textit{GPT-4o} sees modest gains over the vanilla workflow, GPT-4o-mini improves by 10–20\% on average, with some cases reaching 20 points. \textit{Phi-3.5-vision} shows the most notable improvement, exceeding 20\% on average, with rank@N increasing by 26\%. These results highlight the importance of tuning in optimizing \method, ensuring alignment with human values for more robust and valuable outputs.

\subsection{Human Evaluation}

\begin{table}[htbp!]
\centering
\resizebox{0.48\textwidth}{!}{%
\begin{tabular}{l|ccccc|cccc}
\toprule
\textbf{Rating} & \textbf{5} & \textbf{4} & \textbf{3} & \textbf{2} & \textbf{1} & \textbf{Avg} & $\mathbf{\geq 4}$ & $\mathbf{\geq 3}$ & $\mathbf{\leq 2}$ \\
\midrule
Baseline           &  47 & 71 & 92 & 46 & 44 & 3.10 & 118 & 210 & 90 \\
TablePilot (Vanilla) & 114 & 61 & 79 & 25 & 21 & 3.74 & 175 & 254 & 46 \\
TablePilot (Tuned)    & 146 & 75 & 48 & 28 &  3 & \textbf{4.11} & \textbf{221} & \textbf{269} & \textbf{31} \\
\bottomrule
\end{tabular}
}
\caption{Results of human evaluation ratings}
\label{tab:human_evaluation}
\end{table}

\begin{table*}[htbp]
\centering
\resizebox{0.98\textwidth}{!}{%
\begin{tabular}{l|cccccc|c}
\toprule
\multirow{2}{*}{\textbf{Method}} 
& \multicolumn{2}{c}{\textbf{Basic Analysis}} 
& \multicolumn{2}{c}{\textbf{Data Visualization}} 
& \multicolumn{2}{c|}{\textbf{Statistics Modeling}}
& \multirow{2}{*}{\textbf{Overall Recall@N}} \\
\cmidrule(lr){2-3}\cmidrule(lr){4-5}\cmidrule(lr){6-7}
& \textbf{ExecRate} & \textbf{Recall@N}
& \textbf{ExecRate} & \textbf{Recall@N}
& \textbf{ExecRate} & \textbf{Recall@N}
& \\
\midrule

\rowcolor{gray!20}
\textbf{Vanilla}
& 99.67
& \textbf{50.67}
& \textbf{99.67}
& \textbf{69.37}
& 99.44
& 59.58
& \textbf{87.67}
\\
\midrule

w/o sampling
& 98.04 \textcolor{gray}{(-1.63)}
& 48.67 \textcolor{gray}{(-2.00)}
& 98.20 \textcolor{gray}{(-1.47)}
& 65.31 \textcolor{gray}{(-4.06)}
& 98.53 \textcolor{gray}{(-0.91)}
& 58.75 \textcolor{gray}{(-0.83)}
& 86.00 \textcolor{gray}{(-1.67)}
\\

w/o sampling \& revision
& 93.27 \textcolor{gray}{(-6.40)}
& 39.00 \textcolor{gray}{(-11.67)}
& 93.20 \textcolor{gray}{(-6.47)}
& 63.61 \textcolor{gray}{(-5.76)}
& 86.62 \textcolor{gray}{(-12.82)}
& 53.75 \textcolor{gray}{(-5.83)}
& 82.00 \textcolor{gray}{(-5.67)}
\\

w/o explanation
& \textbf{99.93} \textcolor{gray}{(+0.26)}
& 46.00 \textcolor{gray}{(-4.67)}
& 99.27 \textcolor{gray}{(-0.40)}
& 63.61 \textcolor{gray}{(-5.76)}
& \textbf{99.56} \textcolor{gray}{(+0.12)}
& \textbf{62.08} \textcolor{gray}{(+2.50)}
& 83.67 \textcolor{gray}{(-4.00)}
\\

w/o explanation \& revision
& 99.33 \textcolor{gray}{(-0.34)}
& 38.33 \textcolor{gray}{(-12.34)}
& 97.47 \textcolor{gray}{(-2.20)}
& 62.24 \textcolor{gray}{(-7.13)}
& 96.89 \textcolor{gray}{(-2.55)}
& 49.58 \textcolor{gray}{(-10.00)}
& 79.67 \textcolor{gray}{(-8.00)}
\\

w/o sampling \& explanation
& 99.60 \textcolor{gray}{(-0.07)}
& 38.67 \textcolor{gray}{(-12.00)}
& 97.33 \textcolor{gray}{(-2.34)}
& 62.59 \textcolor{gray}{(-6.78)}
& 98.44 \textcolor{gray}{(-1.00)}
& 49.17 \textcolor{gray}{(-10.41)}
& 83.00 \textcolor{gray}{(-4.67)}
\\

w/o all
& 94.73 \textcolor{gray}{(-4.94)}
& 39.67 \textcolor{gray}{(-11.00)}
& 93.87 \textcolor{gray}{(-5.80)}
& 37.76 \textcolor{gray}{(-31.61)}
& 89.19 \textcolor{gray}{(-10.25)}
& 45.83 \textcolor{gray}{(-13.75)}
& 71.67 \textcolor{gray}{(-16.00)}
\\

\bottomrule
\end{tabular}
}
\caption{Impact of removing several components on ExecRate and Recall@N across different tasks (all values in \%)}
\label{tab:ablation}
\end{table*}
\begin{table*}[htbp]
\centering
\resizebox{0.98\linewidth}{!}{%
\begin{tabular}{l|cccccc|cc}
\toprule
\textbf{Method} 
& \multicolumn{2}{c}{\textbf{Basic Analysis}} 
& \multicolumn{2}{c}{\textbf{Data Visualization}} 
& \multicolumn{2}{c}{\textbf{Statistics Modeling}}
& \multicolumn{2}{|c}{\textbf{Overall}} \\
\cmidrule(lr){2-3}\cmidrule(lr){4-5}\cmidrule(lr){6-7}\cmidrule(lr){8-9}
& \textbf{Recall@5} & \textbf{Recall@3}
& \textbf{Recall@5} & \textbf{Recall@3}
& \textbf{Recall@5} & \textbf{Recall@3}
& \textbf{Recall@5} & \textbf{Recall@3} \\
\midrule
\textbf{ranking}
& 21.07 & 14.05
& 48.81 & 35.84
& 28.91 & 15.48
& 70.90 & 53.51 \\
\textbf{w/o ranking} 
& 16.67 \textcolor{gray}{(-4.40)} & 11.56 \textcolor{gray}{(-2.49)}
& 39.80 \textcolor{gray}{(-9.01)} & 23.36 \textcolor{gray}{(-12.48)}
& 22.92 \textcolor{gray}{(-5.99)} & 15.00 \textcolor{gray}{(-0.48)}
& 57.33 \textcolor{gray}{(-13.57)} & 40.22 \textcolor{gray}{(-13.29)} \\
\bottomrule
\end{tabular}
}
\caption{Impact of removing ranking on Recall@K across different tasks (all values in \%)}
\label{tab:rank}
\end{table*}

Automatic quantitative evaluation of tabular data analysis recommendations has many limitations, as it cannot fully assess the concrete quality of the recommended analysis results. To further assess the generation quality of our framework, we conducted a human evaluation comparing the baseline model, \method (Vanilla), and \method (Tuned). Given that the ranking module of \method cannot fully capture certain qualitative aspects, we designed evaluation criteria requiring human judgment. These criteria include Practicality, measuring the relevance and utility of recommended operations in real-world data analysis; Clarity, assessing query unambiguity and result interpretability; and Insightfulness, evaluating the extent to which generated results provide meaningful and non-trivial insights. Five experienced data analysts conducted the evaluation under randomized blind selection protocols, assigning scores based on predefined criteria. Each dimension was rated on a 1–5 scale, with higher scores indicating superior performance.

We analyzed the score distribution across the three frameworks on 300 test tables from the \textbackslash dataset. As shown in Table~\ref{tab:human_evaluation}, \method (Tuned) consistently outperformed other models, achieving the highest mean score, the largest proportion of high-rated outputs (ratings = 5, $\geq$ 4, and $\geq$ 3), and the lowest proportion of low-rated outputs (rating $\leq$ 2). To validate statistical significance, we applied the Wilcoxon signed-rank test \cite{wilcoxon1963critical}, a non-parametric hypothesis test for paired comparisons assessing population mean rank differences. At a 95\% confidence level, results confirm significant improvements favoring \method (Tuned) over both the baseline and \method (Vanilla), as well as \method (Vanilla) over the baseline. These findings substantiate that optimizing the ranked list via \alignmethod leads to measurable improvements in the quality of recommended analytical operations.

\subsection{Ablation Study}
The ablation study results are presented in Table~\ref{tab:ablation} and Table~\ref{tab:rank}. In this experiments, we examine the contributions of key components within the \method workflow, specifically assessing the impact of the Table Explanation, Revision, and Ranking modules on the quality of generated data analysis recommendations. The baseline results represent a system without any of these modules.

Experimental results indicate that nearly all designed components contribute to performance improvements in \method. However, some performance drops can also be attributed to the \textbf{diverse analysis trade-off effect}.

\subsection{Analysis of \alignmethod}
Our proposed \alignmethod implemented by DPO has demonstrated robust and consistent improvements across various model configurations and tasks. When applied to these models, this alignment consistently enhances performance. In addition to these performance gains, the alignment plays a crucial role in aligning model outputs with human values. By optimizing for results that resonate with human preferences, \method effectively selects responses that are better matched to human values, thereby increasing the recall of value-aligned outputs. This alignment ensures that the models not only perform better in a technical sense but also generate outputs that are more in line with the ethical and qualitative standards expected by users.

The differences in performance gains observed across various models when integrating \alignmethod can be primarily attributed to variations in model architecture, baseline fine-tuning, and the inherent alignment with human values. Larger models, such as \textit{GPT-4o}, possess a higher capacity for capturing intricate patterns, which may result in relatively modest incremental improvements with DPO, whereas smaller models like \textit{GPT-4o-mini} tend to benefit more significantly. Models that have undergone extensive specialized training might experience smaller relative gains because their baseline performance is already near optimal, while others with less refined pre-training show a more pronounced boost when DPO is applied.

\section{Conclusion}

In this paper, we introduce \method, a comprehensive data analysis recommendation framework powered by large language models. Extensive experiments demonstrate \method’s superior performance, marking a new milestone in tabular data analysis recommendation.


\section*{Limitations}
Our work is an exploratory study on comprehensive tabular data analysis, with several limitations and areas for improvement, such as better data curation, advanced multi-modal training, and efficiency optimization, etc. For further discussion on the extendability of \method and future work, please refer to Appendix~\ref{ap:future}.


\bibliography{custom}

\appendix
\clearpage

\onecolumn
\DoToC

\clearpage
\twocolumn

\section{Extendability Analysis and Future Works}
\label{ap:future}
In this paper, we present an exploratory study on comprehensive tabular data analysis. Several important extensions of our proposed framework, \method, remain open for future work. \\

\noindent \textbf{Data Curation.}
We provide the dataset \dataset to support model training and to validate the performance of \method. However, the current dataset has several limitations: it is relatively small in scale, lacks image-contrastive data necessary for effective multi-modal SFT and DPO, and contains limited high-quality samples. We believe that with more carefully curated data and improved data construction pipelines, \method could achieve significantly better performance and enable more powerful analytical capabilities. \\

\noindent \textbf{Multi-Modal Training.}
One significant direction for extending \method lies in the integration of multi-modal GPT-based models, such as multi-modal SFT and DPO. As previously mentioned, higher-quality multi-modal training data is crucial for achieving better performance. In addition, current GPT-series models on the Azure platform do not yet support multi-modal DPO, limiting our ability to fully leverage visual information during optimization. Multi-modal DPO could substantially improve \method’s ability to evaluate and analyze results based on figures and visualizations. Furthermore, how to design multi-stage training pipelines that balance SFT and DPO to achieve optimal model performance remains an open challenge. We believe that, with the integration of more advanced multi-modal capabilities, \method can generate richer analytical insights, enhance contextual understanding, and better align with how human analysts interpret complex, heterogeneous data sources. \\

\noindent \textbf{Analysis Modularization.}
The current version of \method supports three types of analysis: Basic Analysis, Table Visualization, and Statistical Modeling. These analyses are implemented in a modularized manner, allowing flexible composition and extension. As these three represent some of the most classical forms of tabular data analysis, they provide a strong foundation for various use cases. In the future, more diverse or specialized analysis modules can be easily integrated into \method, showcasing the flexibility of our framework. Furthermore, in different downstream application scenarios, \method can adaptively select and combine specialized analysis modules to better address domain-specific needs. \\

\noindent \textbf{System Internal Interaction.}
The current framework of \method is unidirectional, with different analysis modules operating in parallel without internal interaction. In the future, we aim to extend \method into a multi-agent system, enabling richer interactions between modules. For example, different analysis modules could complement and enrich each other’s data, and the ranking module could provide feedback to guide the analysis modules. We believe that such a design would make the system more intelligent and capable of generating higher-quality analytical recommendations. \\

\noindent \textbf{Efficiency Optimization.}
Our current \method framework involves multiple large language model (LLM) calls, which can lead to efficiency issues. In the future, we plan to improve efficiency by replacing certain modules with smaller language models or well-trained traditional machine learning models. Additionally, optimizing and compressing prompts will help streamline the pipeline and further enhance overall efficiency.

\section{Evaluation Metrics}
\label{ap:metric}

In our experiments, we adopt two primary metrics to evaluate system performance comprehensively: \textit{Execution Rate} (abbreviated as \textit{ExecRate}) and \textit{Recall}.

\textit{ExecRate} quantifies the accuracy and stability of generated code by measuring whether it executes without error and returns the expected output. This metric is consistently applied across all modules (Basic Analysis, Table Visualization, and Statistical Modeling) by calculating the ratio of successful executions to the total number of generated outputs.

\textit{Recall} serves as our key indicator for retrieval accuracy, assessing whether the correct result appears among the top-ranked candidates. We distinguish among three variants: \textit{Recall@All}, \textit{Recall@5}, and \textit{Recall@3}. \textit{Recall@All} checks if the correct result is present anywhere in the candidate set, while \textit{Recall@5} and \textit{Recall@3} evaluate if it ranks within the top five and top three candidates, respectively. For Basic Analysis, success is defined by an exact match of the output table to the expected result. In Table Visualization, the evaluation hinges on the precise match of generated chart information—including x\_fields, y\_fields, and chart\_type. For Statistical Modeling, evaluation is further subdivided into Regression, Correlation, and Forecast tasks. Specifically, Regression is deemed successful if the \textit{Mean Absolute Percentage Error} (\textit{MAPE}) is $\leq 1.0$, Correlation if the \textit{p-value} is $< 0.05$, and Forecast if the \textit{R-squared value} is $> 0.9$, with the additional requirement that the table column relationships align with the expected structure. These metrics  ensures a robust assessment of both execution reliability and the system’s ability to prioritize accurate results.



\section{Dataset details}
\label{ap:dataset}

\begin{figure}
    \centering
    \includegraphics[width=0.8\linewidth]{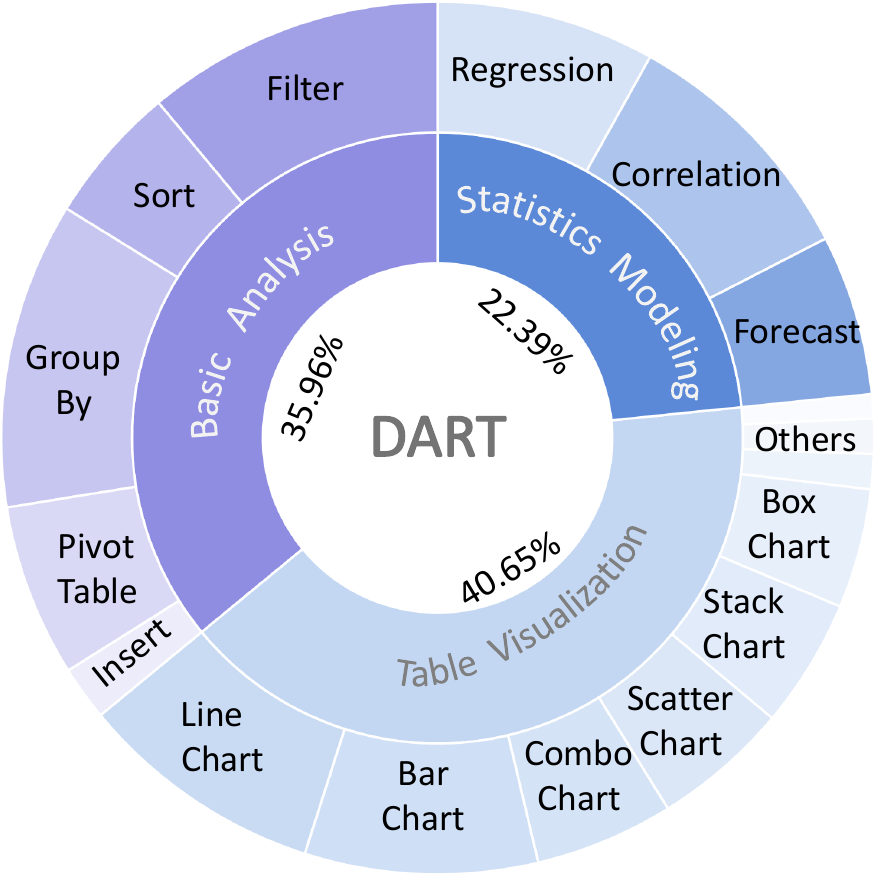}
    \caption{Statistics of the test split of the \dataset dataset. We can categorize data analysis tasks into three major groups: Basic Analysis (35.96\%), Table Visualization (40.65\%), and Statistics Modeling (22.39\%). This distribution highlights the diversity of analytical tasks covered in the dataset.}
    \label{fig:dataset_stat}
\end{figure}

To support and validate the performance of \method, we conducted an investigation on the table dataset \dataset. Existing datasets, such as those in the Text2SQL domain \cite{xu2018sql, lei2024spider}, which focus on SQL-like analytical QA tasks, and the Table2Charts domain \cite{han2023chartllama, zhou2021table2charts}, which specializes in table-to-chart QA and conversion, are designed for specific domains rather than comprehensive analysis. Additionally, even common analysis datasets like Text2Analysis \cite{zhou2020table2analysis} are primarily designed for TableQA scenarios, making them misaligned with our proposed task of zero-turn data analysis recommendation. As a result, we constructed a custom dataset to better support our tasks. To the best of our knowledge, \dataset is the first dataset dedicated to recommending comprehensive tabular data analysis operations.

Our dataset construction process was inspired by Table2Charts \cite{zhou2021table2charts}, which contains a collection of real-world tables. We leveraged these tables as a foundation for synthetic data generation, ensuring that the dataset retained realistic tabular structures while expanding its applicability to our target tasks. The data generation process was conducted using \textit{o1} and consisted of three main step:
\\

\begin{enumerate}[leftmargin=*, itemsep=0pt, labelsep=5pt, topsep=0pt]
    \item \textbf{Table Selection.} We filtered the tables from Table2Charts, selecting those that were most suitable for data analysis tasks with strong tabular structures. This selection process ensured that the tables contained sufficient variability in structure, numerical and categorical distributions, and contextual relevance for analytical queries. 
    \item \textbf{Label Generation.} For each of the three tasks, \textit{o1} was employed to generate a diverse set of queries and their corresponding code implementations. The queries were designed to cover a range of complexity levels, from simple transformations to advanced statistical modeling tasks. The code snippets were generated in Python, incorporating libraries such as \texttt{Pandas}, \texttt{Matplotlib}, and \texttt{StatsModels}, ensuring their practical applicability. However, from all the generated queries and code, we carefully selected only those that were able to successfully produce the expected results.
    \item \textbf{Human Evaluation.} We manually curated a subset of 300 tables to ensure diversity in structure and analytical needs. From the generated triplets, we selected those that met specific criteria for clarity, correctness, and so on based on human preferences. This process resulted in a test set that reflects real-world analytical tasks. The test set was then used to evaluate the model’s performance, particularly through metric recall, providing a robust benchmark for \method’s capabilities. Finally, \dataset consists of 300 data from different tables. The dataset distribution is shown in Figure~\ref{fig:dataset_stat}.
\end{enumerate}

\section{Detailed Experiment Settings}
\label{ap:settings}

\begin{table*}[ht]
\centering
\resizebox{\textwidth}{!}{%
\begin{tabular}{@{}llcc@{}}
\toprule
\textbf{Model}                         & \textbf{Parameter}              & \textbf{Supervised Fine-Tuning (SFT)} & \textbf{Direct Preference Optimization (DPO)} \\ \midrule
\multirow{3}{*}{GPT-4o / GPT-4o-mini}     & Learning Rate                  & $1 \times 10^{-6}$                    & $5 \times 10^{-7}$                           \\ \cline{2-4} 
                                        & Number of Epochs               & 6                                      & 2                                           \\ \cline{2-4} 
                                        & Batch Size                     & 64                                     & 32                                          \\ \midrule
\multirow{4}{*}{Phi-3.5-Vision} & Learning Rate                  & $1 \times 10^{-5}$                    & $5 \times 10^{-7}$                           \\ \cline{2-4} 
                                        & Number of Epochs               & 3                                      & 2                                           \\ \cline{2-4} 
                                        & Batch Size                     & 8                                      & 8                                           \\ \cline{2-4} 
                                        & Full-Parameter Training        & Yes                                    & No                                          \\ \bottomrule
\end{tabular}
}
\caption{Training Parameters for GPT-4o, GPT-4o-mini, and Phi-3.5-Vision Models}
\label{tab:training}
\end{table*}

We use the \texttt{trl} package to fine-tune open-source models on a workstation equipped with 4 $\times$ A100 Nvidia GPUs. LoRA fine-tuning \cite{hu2021lora} is applied to train GPT-series models on the Azure platform \footnote{https://azure.microsoft.com/en-us/}, leveraging its scalable infrastructure. The models used in our experiments include \textit{GPT-4o} (gpt-4o-08-06), \textit{GPT-4o-mini} (gpt-4o-mini-2024-07-18), and \textit{Phi-3.5-Vision} (microsoft/Phi-3.5-vision-instruct). OpenAI \textit{o1} used in our study are \textit{o1-2024-12-17}. The detailed training parameters can be found in Table~\ref{tab:training}.

For inference, the parameters are set as follows for all models, including both open-source and private models: temperature is 0, max tokens is 6000, top-p is 0.95, frequency penalty is 0, presence penalty is 0, and stop is set to None. All other settings are configured to their default values. Inference stage is also conducted in 4 $\times$ A100 Nvidia GPUs.

In the SFT phase, we used \textit{o1} to generate a batch of data tailored to the task requirements. To ensure the quality of the data, we employed LLM-based evaluation along with manual sampling. For fine-tuning the module-based analysis, we used 800 training samples and 100 validation samples for both the basic analysis and table visualization modules. Due to the complexity of its tasks, the statistics modeling module was trained using 1,100 samples, with 100 samples reserved for validation.

In the DPO phase, we first performed an SFT run on the ranking module using 342 ranked samples generated by \textit{o1}. Afterward, DPO training was conducted with 1,000 positive and negative samples. The positive samples consisted of ranking results generated by \textit{o1}, which were manually adjusted based on preference calibration. The negative samples were disordered ranking results produced by \textit{GPT-4o-mini}.

\begin{table*}[htbp]
\centering
\resizebox{0.98\textwidth}{!}{%
\begin{tabular}{ll|cccccc|cc}
\toprule
\textbf{Analysis} & \textbf{Phi-3.5-vision Rank} 
& \multicolumn{2}{c}{\textbf{Basic Analysis}} 
& \multicolumn{2}{c}{\textbf{Table Visualization}} 
& \multicolumn{2}{c}{\textbf{Statistics Modeling}} 
& \multicolumn{2}{c}{\textbf{Overall}} \\
\cmidrule(lr){3-4}\cmidrule(lr){5-6}\cmidrule(lr){7-8}\cmidrule(lr){9-10}
& & Recall@3 & Recall@5 
& Recall@3 & Recall@5 
& Recall@3 & Recall@5 
& Recall@3 & Recall@5 \\
\midrule
\multirow{4}{*}{\textbf{Phi-3.5-vision}} 
& Rank SFT
& 6.85 & 14.04 
& 2.79 & 4.18 
& 15.88 & 22.75 
& 20.89 & 32.19 \\
& Rank SFT-V
& 5.14 \textcolor{langred}{(-1.71)} & 13.01 \textcolor{langred}{(-1.03)}
& 1.74 \textcolor{langred}{(-1.05)} & 3.14 \textcolor{langred}{(-1.04)}
& 19.31 \textcolor{langgreen}{(+3.43)} & 21.89 \textcolor{langred}{(-0.86)}
& 21.23 \textcolor{langgreen}{(+0.34)} & 30.14 \textcolor{langred}{(-2.05)} \\
& Rank SFT \& DPO
& 8.90 & 15.07 
& 1.74 & 3.83 
& 18.88 & 23.61 
& 23.97 & 32.88 \\
& Rank SFT-V \& DPO
& 7.53 \textcolor{langred}{(-1.37)} & 14.38 \textcolor{langred}{(-0.69)}
& 1.74 \textcolor{gray}{(0.00)} & 2.09 \textcolor{langred}{(-1.74)}
& 19.31 \textcolor{langgreen}{(+0.43)} & 25.32 \textcolor{langgreen}{(+1.71)}
& 23.63 \textcolor{langred}{(-0.34)} & 32.19 \textcolor{langred}{(-0.69)} \\
\midrule
\multirow{4}{*}{\textbf{GPT-4o}} 
& Rank SFT 
& 14.67 & 24.00 
& 20.07 & 28.57 
& 13.75 & 20.00 
& 39.57 & 52.00 \\
& Rank SFT-V
& 10.76 \textcolor{langred}{(-3.91)} & 20.33 \textcolor{langred}{(-3.67)}
& 26.87 \textcolor{langgreen}{(+6.80)} & 35.71 \textcolor{langgreen}{(+7.14)}
& 13.75 \textcolor{gray}{(0.00)} & 23.75 \textcolor{langgreen}{(+3.75)}
& 42.00 \textcolor{langgreen}{(+2.43)} & 55.33 \textcolor{langgreen}{(+3.33)} \\
& Rank SFT \& DPO
& 12.67 & 23.00 
& 21.77 & 30.27 
& 12.08 & 20.42 
& 38.33 & 53.33 \\
& Rank SFT-V \& DPO
& 17.00 \textcolor{langgreen}{(+4.33)} & 25.33 \textcolor{langgreen}{(+2.33)}
& 27.21 \textcolor{langgreen}{(+5.44)} & 38.10 \textcolor{langgreen}{(+7.83)}
& 13.33 \textcolor{langgreen}{(+1.25)} & 18.75 \textcolor{langred}{(-1.67)}
& 46.00 \textcolor{langgreen}{(+7.67)} & 60.00 \textcolor{langgreen}{(+6.67)} \\
\bottomrule
\end{tabular}
}
\caption{Performance on Recall@3 and Recall@5 with different Phi-3.5-vision Rank}
\label{tab:phi35_rank_validation}
\end{table*}

\section{Ranking Criteria}
\label{ap:criteria}
\method employs a structured prompt with explicit criteria to filter and rank data analysis recommendations using an LLM. The core ranking criteria include:

\begin{enumerate}[leftmargin=*, itemsep=0pt, labelsep=5pt, topsep=0pt]
\item \textbf{Meaningfulness}: Recommendations must offer impactful, insightful queries rather than trivial data representations. Queries should directly facilitate actionable insights.

\item \textbf{Relevance}: Recommendations must align closely with the Table Theme, ensuring analytical coherence with the dataset's core objective.

\item \textbf{Logical Coherence}: Recommendations must follow fundamental data analysis principles, accurately reflecting logical relationships and dataset characteristics.

\item \textbf{Diversity}: Ensures a broad coverage of analytical tasks across basic operations, data visualization, and advanced analyses to maximize insight comprehensiveness.

\item \textbf{Interpretability}: Recommendations should be clear, concise, and easily implementable by analysts without ambiguity.

\item \textbf{Insightfulness}: Prioritizes queries revealing non-obvious patterns, trends, and relationships that significantly enhance understanding of the data.
\end{enumerate}

Additional task-specific constraints are applied to further refine the recommendations, eliminating redundancy, trivial analyses, and logically unsound operations. This structured ranking criteria, embedded within a unified prompt and processed through an LLM, ensures the efficient selection and prioritization of high-quality analytical queries that align with professional analytical standards.

\section{Direct Preference Optimization}
\label{ap:dpo}
Direct Preference Optimization (DPO) \cite{rafailov2023direct} is a reinforcement learning-free approach for fine-tuning large language models (LLMs) using human preferences. Given preference-labeled data pairs \(\{(x, y^+, y^-)\}\), where \(y^+\) is the preferred response and \(y^-\) is the less preferred response for input \(x\), DPO optimizes the policy \(\pi_{\theta}(y | x)\) by maximizing the implicit reward function derived from the Bradley-Terry model:  

\[
r_{\theta}(x, y^+) - r_{\theta}(x, y^-) = \log \frac{\pi_{\theta}(y^+ | x)}{\pi_{\theta}(y^- | x)}
\]

The loss function for DPO is formulated as:

\[
\mathcal{L}(\theta) = - \mathbb{E}_{(x, y^+, y^-)} \left[ \log \sigma \left( \beta \log \frac{\pi_{\theta}(y^+ | x)}{\pi_{\theta}(y^- | x)} \right) \right]
\]

where \(\sigma\) is the sigmoid function and \(\beta\) is a scaling factor that controls the sharpness of preference separation. This formulation ensures that the model directly optimizes preference probabilities while maintaining policy stability and avoiding the high variance introduced by reinforcement learning methods.  

In \alignmethod, we specifically integrate Direct Preference Optimization (DPO) within the ranking module to align data analysis recommendations with human analytical preferences. By assigning higher scores to operations that effectively capture user intent and generate actionable insights, and lower scores to less useful analyses, DPO effectively reinforces outputs aligned with analyst expectations.

This targeted integration of DPO significantly enhances the quality and practical applicability of generated analyses by ensuring accurate alignment with human analytical preferences.


\section{Analysis of Incorporating Vision in Training}

Incorporating vision into the training process proves both valuable and effective. For GPT-4o and GPT-4o-mini, the addition of vision capabilities significantly enhances the ranking module. Compared to pure text-based ranking, these models show improved recall@3 and recall@5 metrics. Specifically, in the Table Visualization Task, GPT-4o-mini demonstrates a 9\% increase in recall@5 and a 12\% increase in recall@3, which contributes substantially to the overall improvements of 5\% in recall@3 and 4\% in recall@5. Due to its smaller scale and comparatively weaker multimodal capabilities relative to GPT-4o, GPT-4o-mini benefits even more from multimodal training in enhancing its ranking ability.

Conversely, Phi-3.5-vision does not benefit from multimodal training; in fact, its performance declines. This decline is primarily attributed to the poor quality of table visualizations generated by Phi-3.5. While we trained the ranking model on high-quality ranking data generated by GPT-4o and GPT-4o-mini, which in turn produced abundant high-quality analysis data, Phi-3.5 generated relatively few examples of data with lower quality. This data disparity, coupled with the inherent limitations of Phi-3.5, makes it challenging for the model to effectively learn to rank lower-quality data, ultimately resulting in reduced performance. 

To verify that Phi-3.5-vision indeed learns to rank multimodal triplets after multimodal SFT, we conducted an experiment using GPT-4o–generated triplets as the basis for ranking, as detailed in Table \ref{tab:phi35_rank_validation}. Our evaluation indicates that employing the multimodal SFT–enhanced Phi-3.5-vision as the ranking module yields an overall recall boost of 3\% to 5\%. Furthermore, in multimodal scenarios—particularly in the Table Visualization task—Phi-3.5-vision achieves an average increase of 6.8\% in recall@3 and recall@5. These findings suggest that while Phi-3.5-vision demonstrates robust multimodal ranking capabilities, its overall performance is nevertheless limited by the suboptimal quality of the triplets it generates.

\clearpage

\section{Case Study}
\label{ap:case study}
\begin{figure*}
    \centering
    \includegraphics[width=0.9\linewidth]{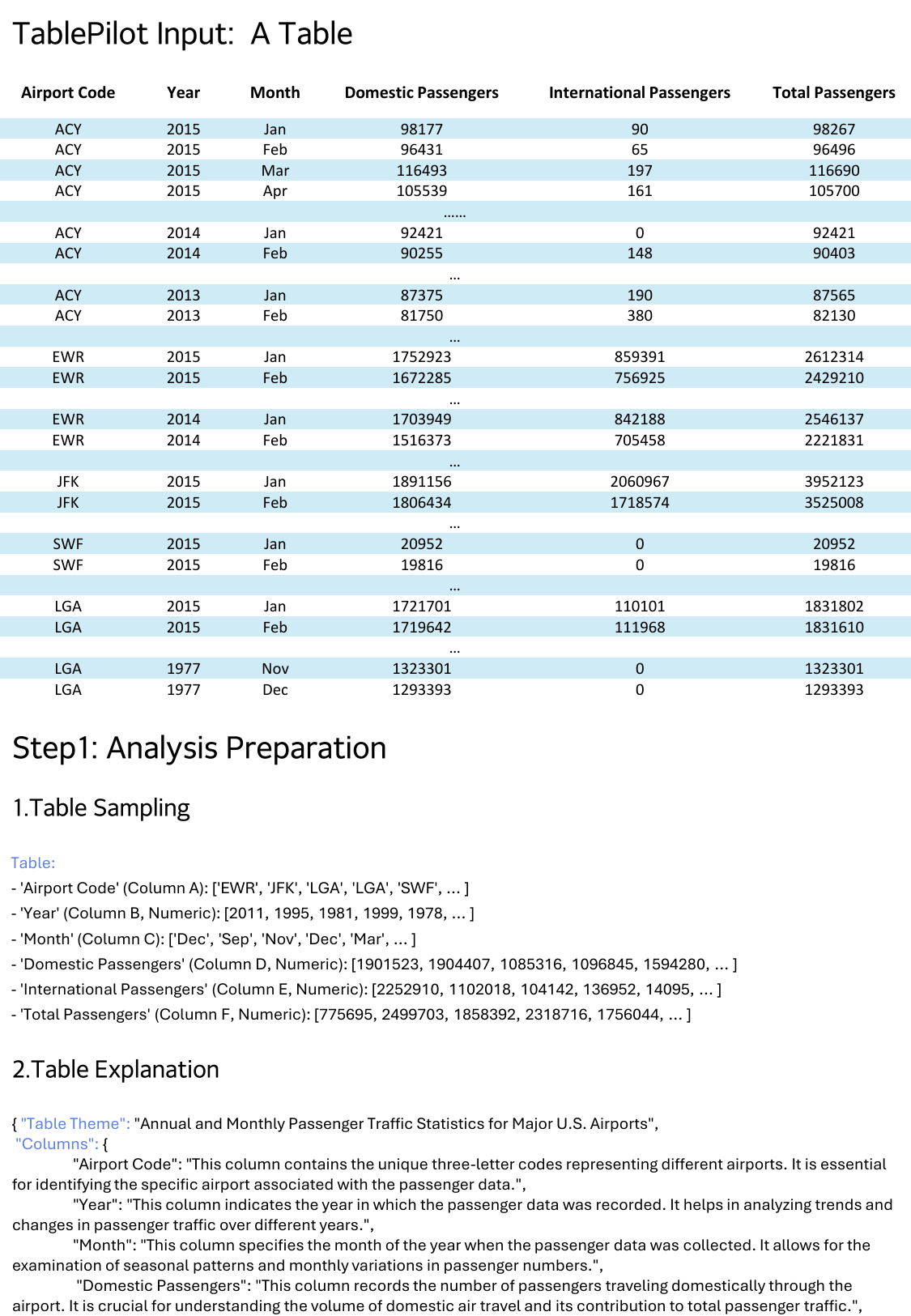}
    \caption{Overview of \method framework case study. Analysis Preparation-1.}
    \label{fig:case_study-1}
\end{figure*}

\begin{figure*}
    \centering
    \includegraphics[width=0.9\linewidth]{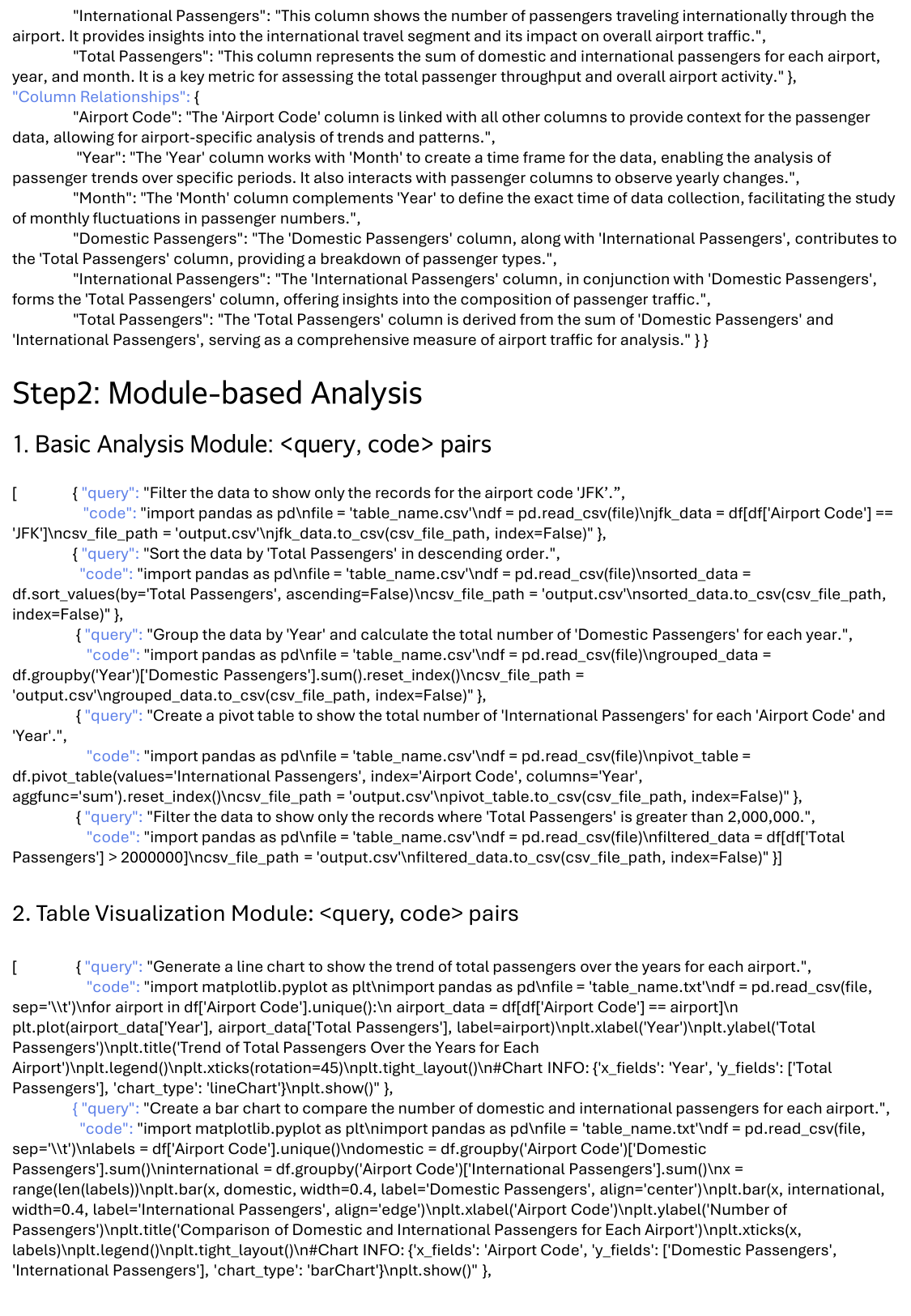}
    \caption{Overview of \method framework case study. Module-based Analysis-1.}
    \label{fig:case_study-2}
\end{figure*}

\begin{figure*}
    \centering
    \includegraphics[width=0.9\linewidth]{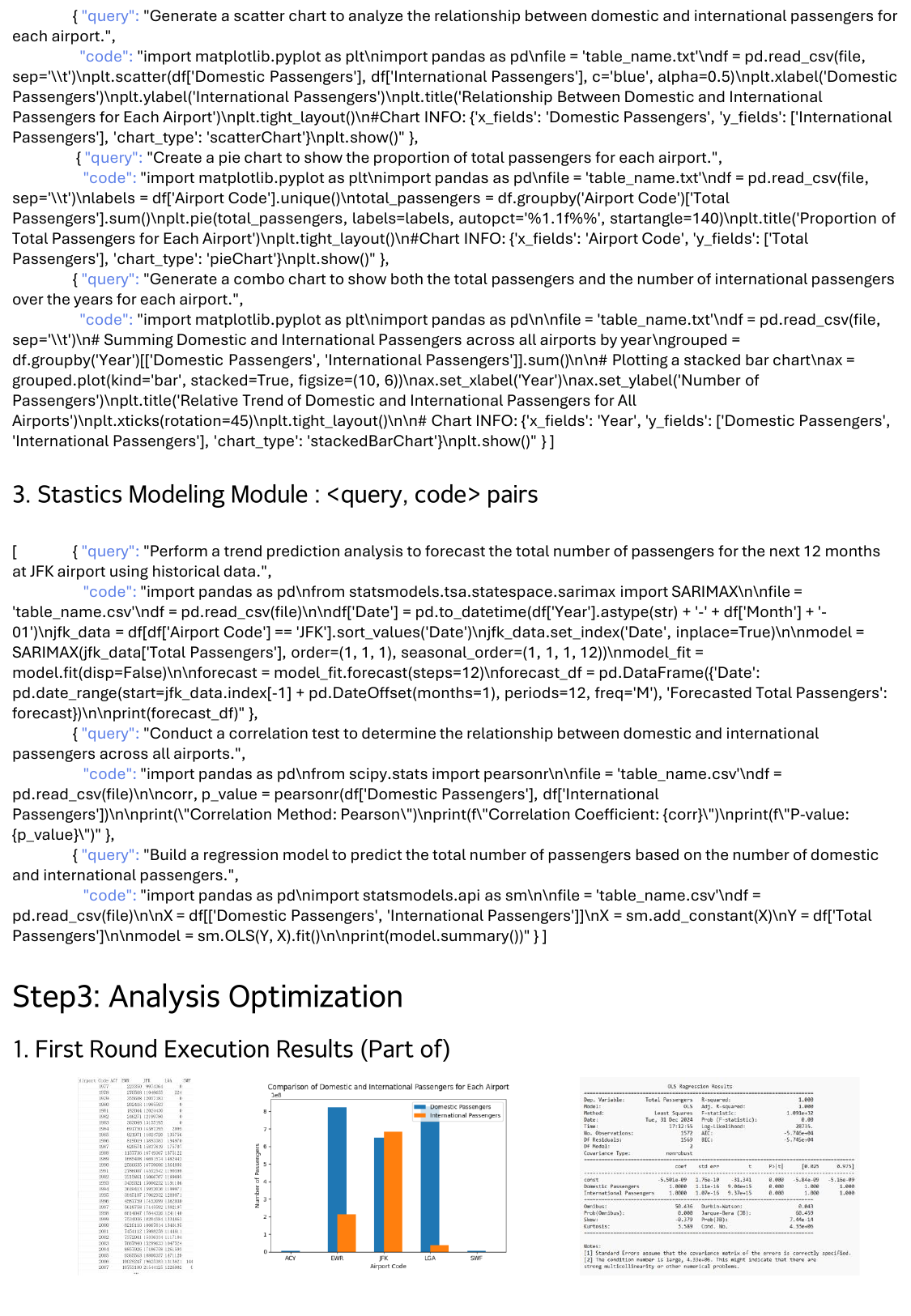}
    \caption{Overview of \method framework case study. Module-based Analysis-2.}
    \label{fig:case_study-3}
\end{figure*}

\begin{figure*}
    \centering
    \includegraphics[width=0.9\linewidth]{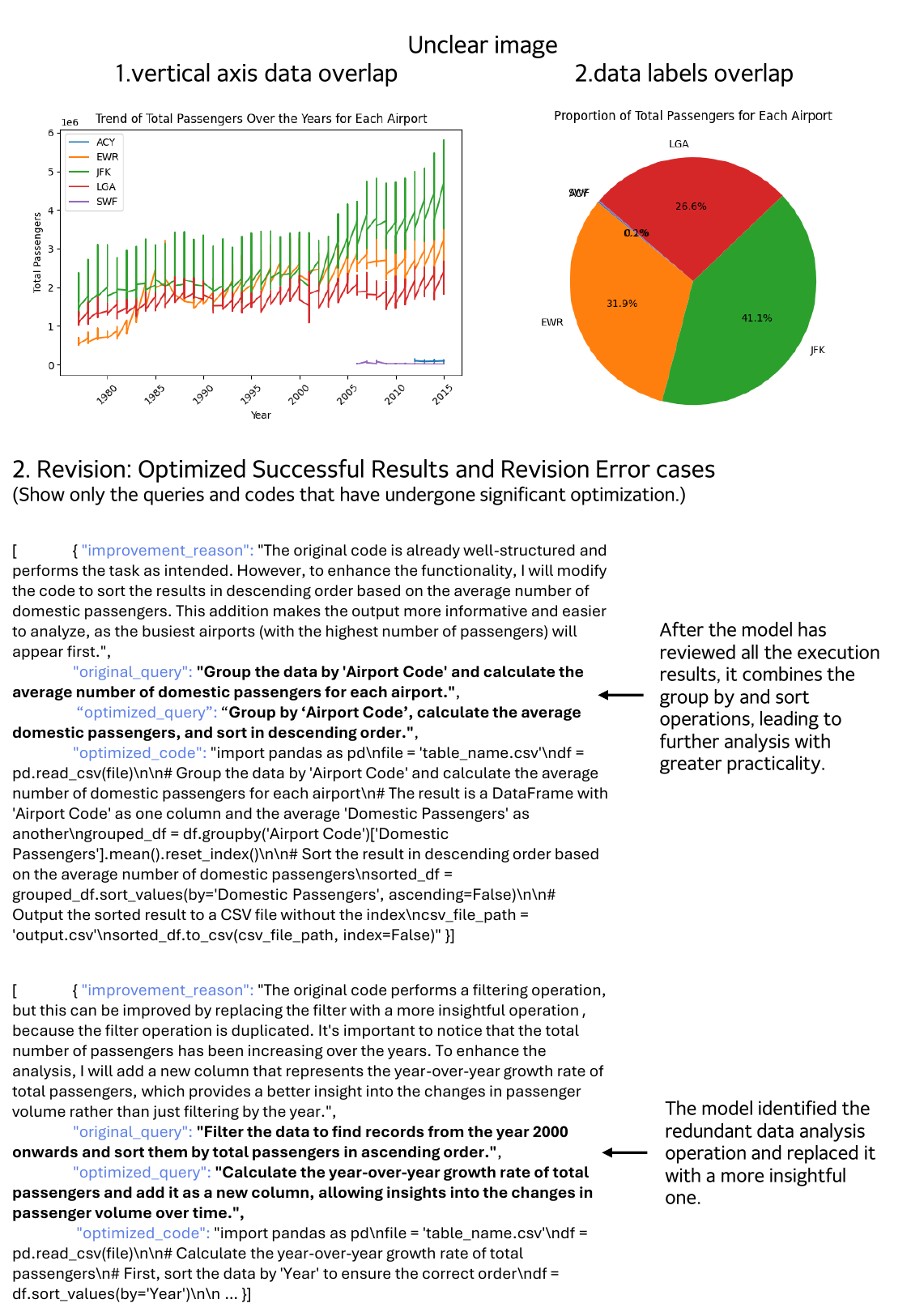}
    \caption{Overview of \method framework case study. Analysis Optimization-1.}
    \label{fig:case_study-4}
\end{figure*}

\begin{figure*}
    \centering
    \includegraphics[width=0.9\linewidth]{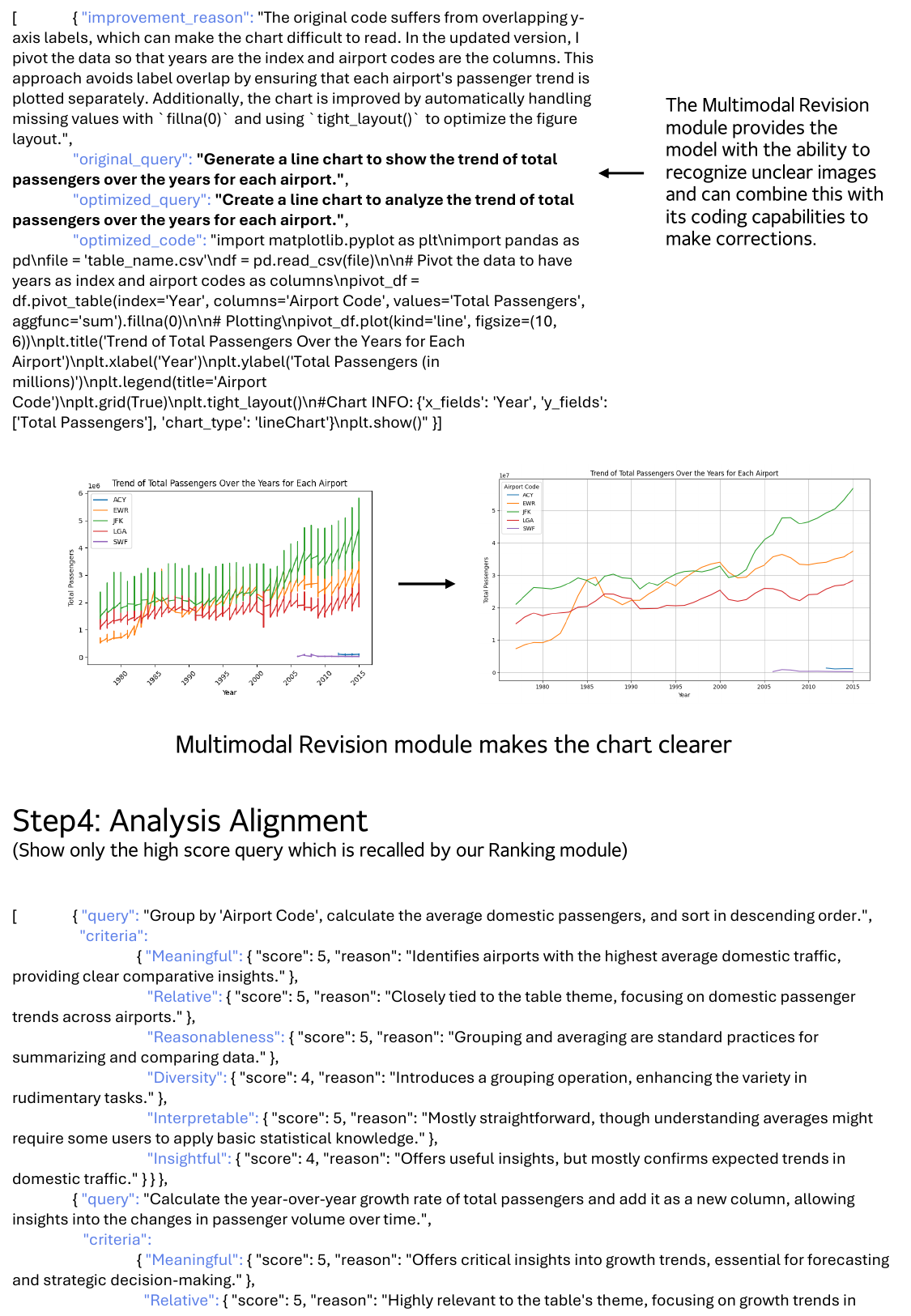}
    \caption{Overview of \method framework case study. Analysis Optimization-2.}
    \label{fig:case_study-5}
\end{figure*}

\begin{figure*}
    \centering
    \includegraphics[width=0.9\linewidth]{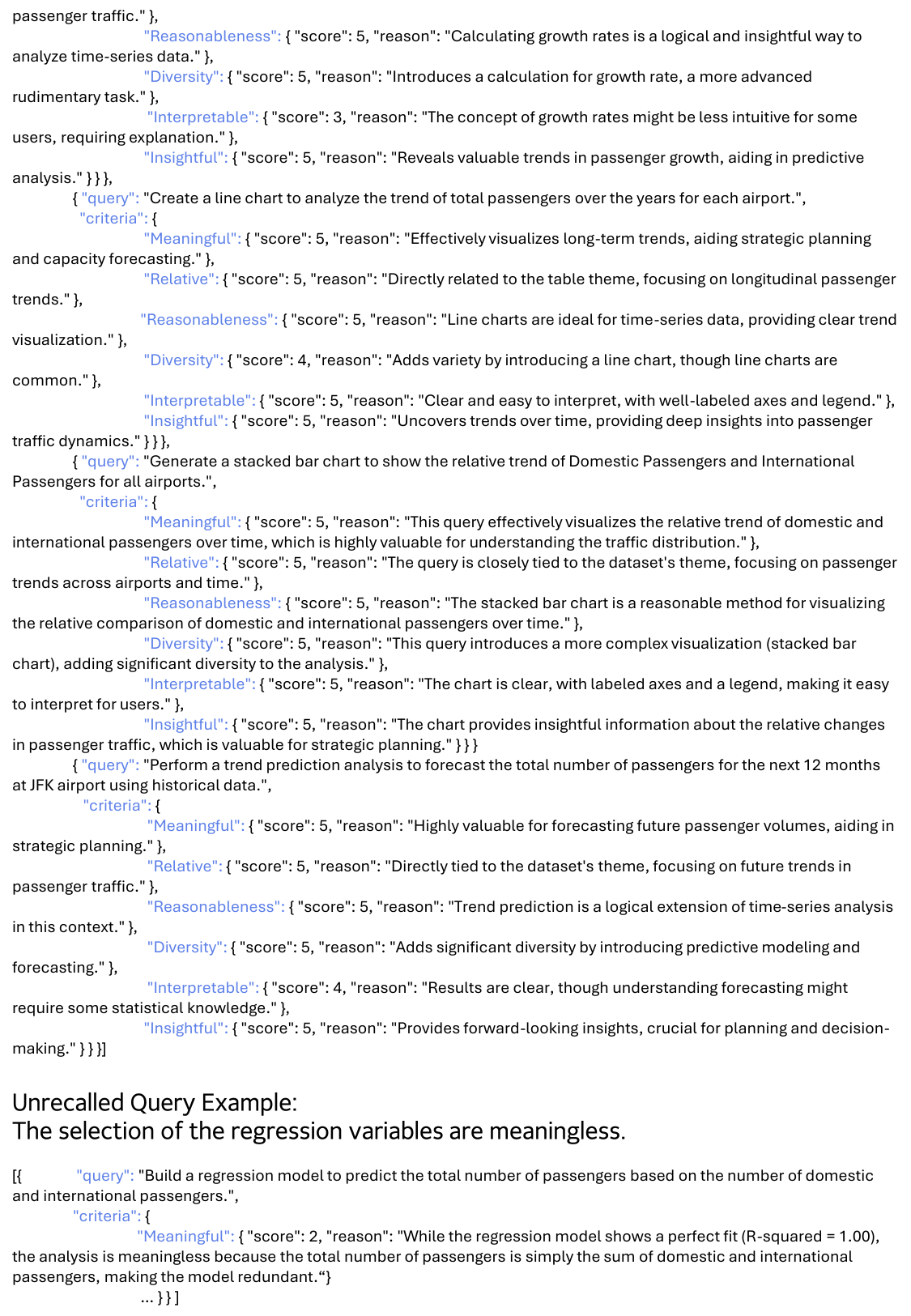}
    \caption{Overview of \method framework case study. Analysis Optimization-3.}
    \label{fig:case_study-6}
\end{figure*}

\begin{figure*}
    \centering
    \includegraphics[width=0.9\linewidth]{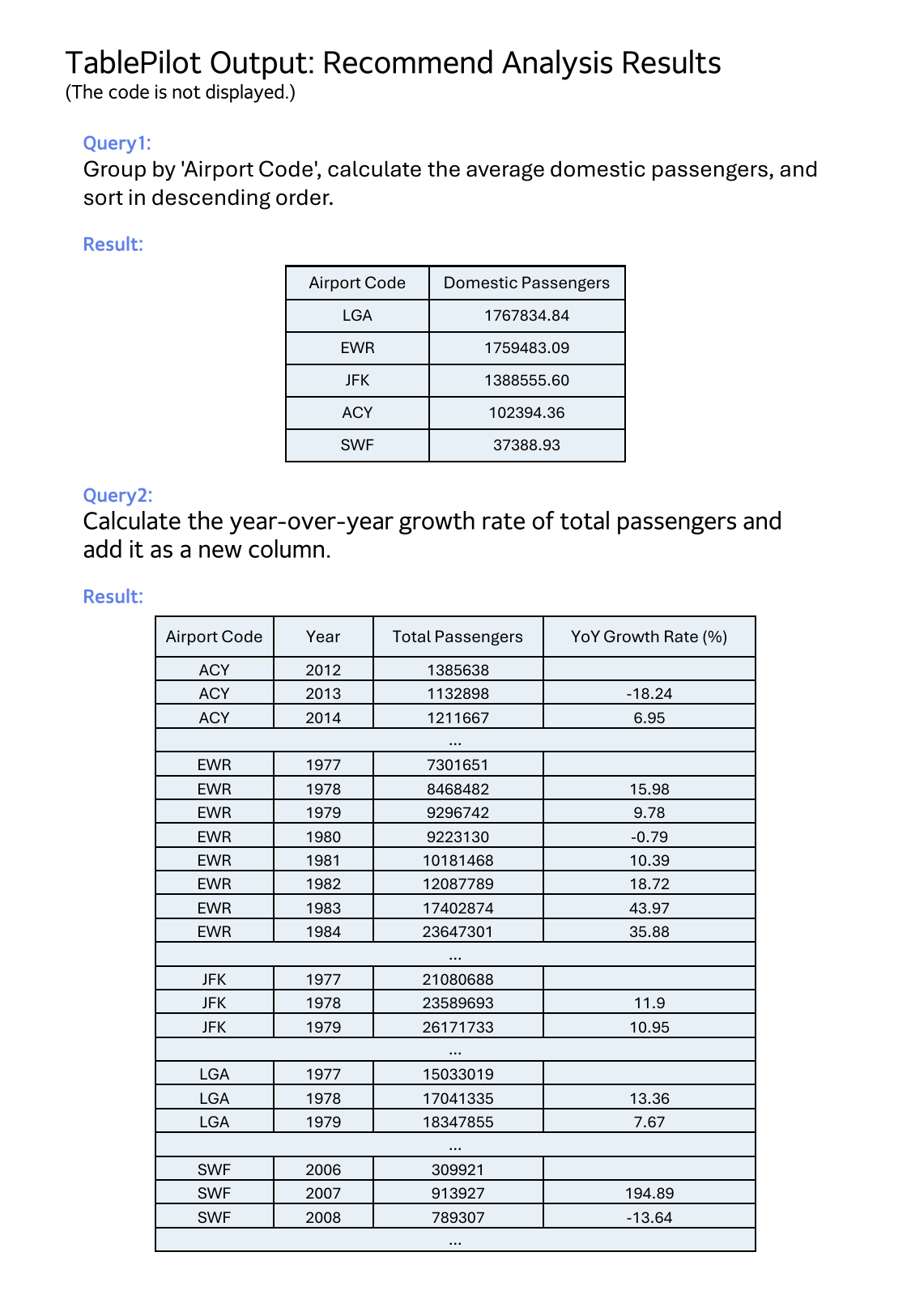}
    \caption{Overview of \method framework case study. \method Output Results-1.}
    \label{fig:case_study-7}
\end{figure*}

\begin{figure*}
    \centering
    \includegraphics[width=0.9\linewidth]{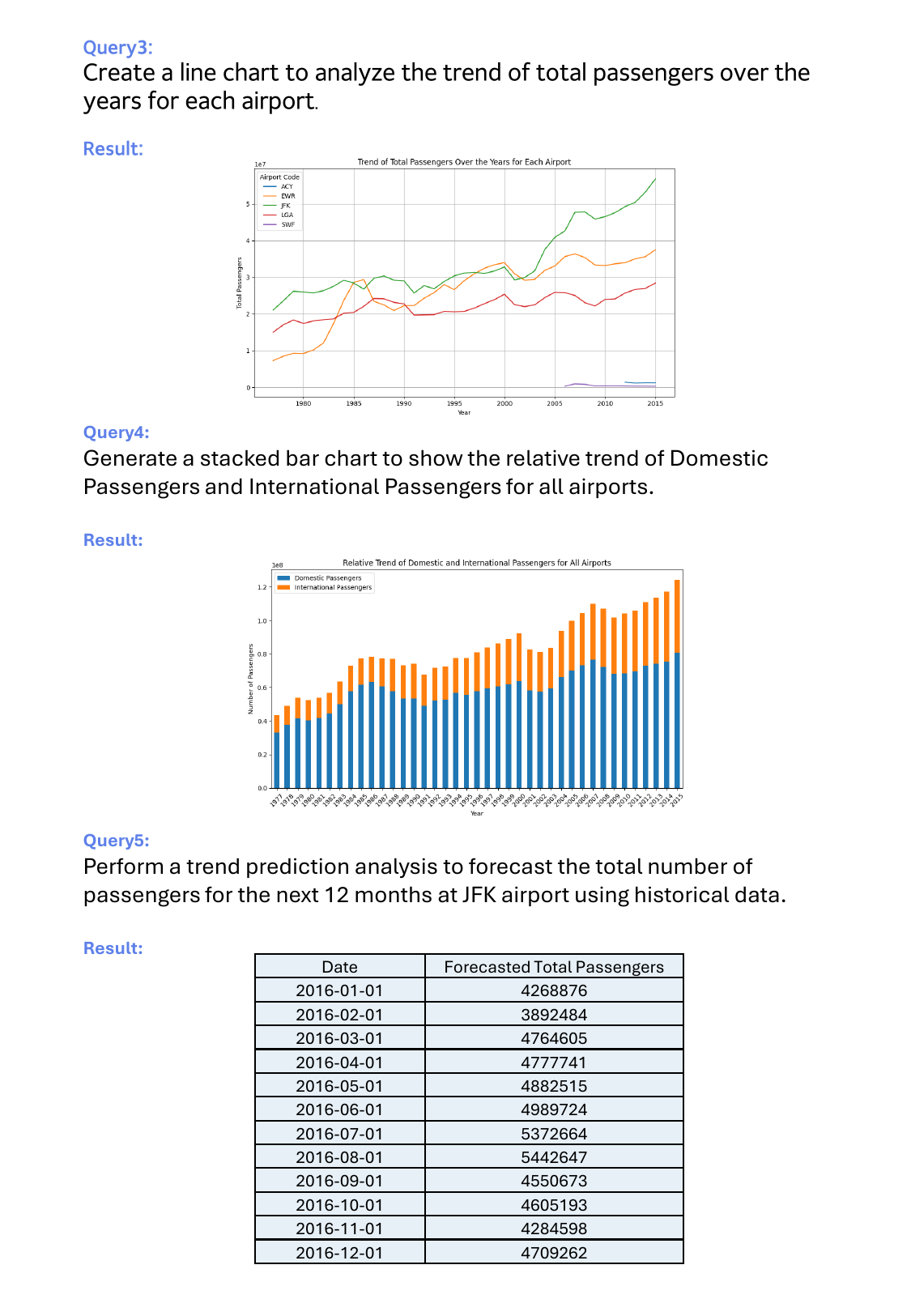}
    \caption{Overview of \method framework case study. \method Output Results-2.}
    \label{fig:case_study-8}
\end{figure*}

Figure~\ref{fig:case_study-1} to Figure~\ref{fig:case_study-8} illustrate a case study demonstrating our \method framework. This case provides a detailed analysis of a real-world example, showcasing the practical applications and effectiveness of \method in generating comprehensive data analysis recommendations.

\clearpage
\section{\method Report Generation}
\label{ap:report}
\begin{figure*}
    \centering
    \includegraphics[width=0.9\linewidth]{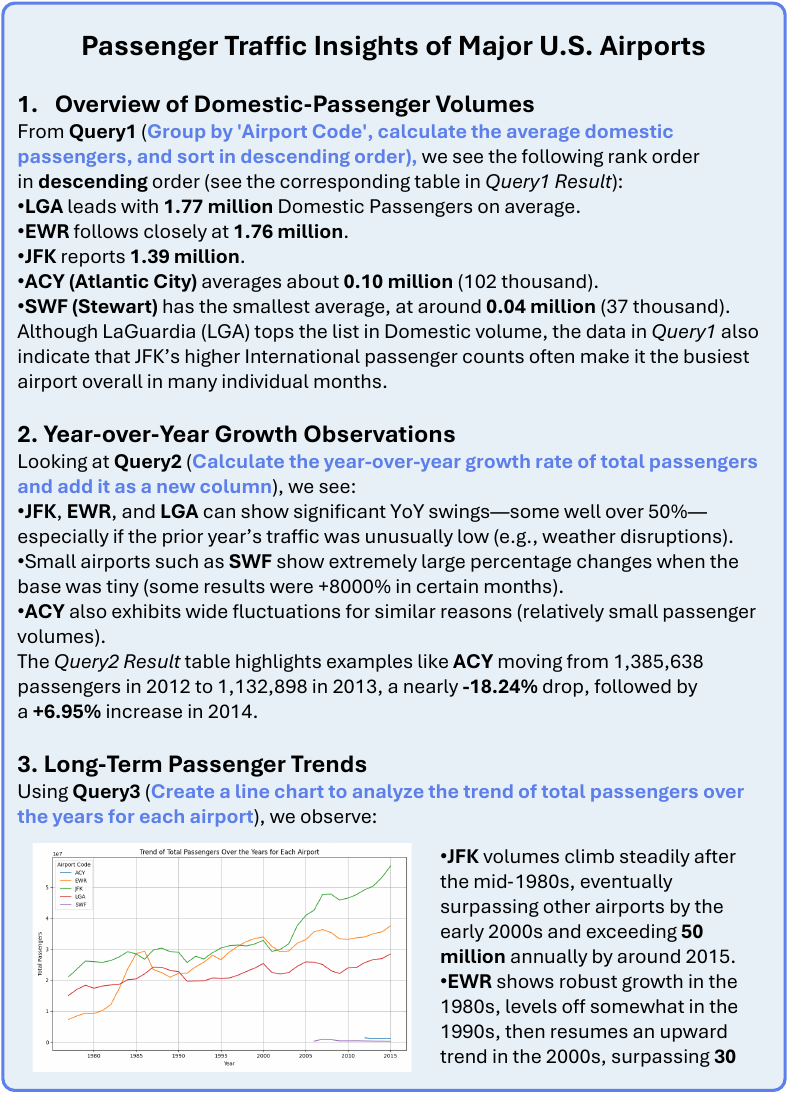}
    \caption{Overview of \method framework report-1.}
    \label{fig:report-1}
\end{figure*}

\begin{figure*}
    \centering
    \includegraphics[width=0.9\linewidth]{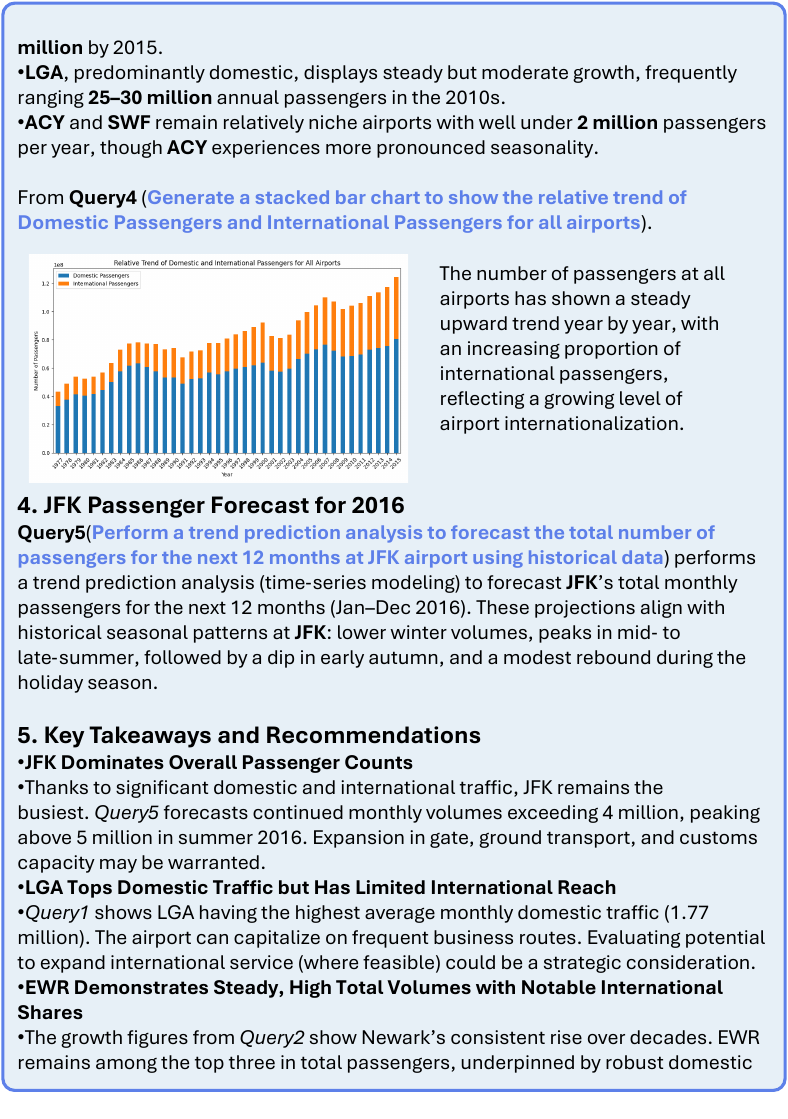}
    \caption{Overview of \method framework report-2.}
    \label{fig:report-2}
\end{figure*}

\begin{figure*}
    \centering
    \includegraphics[width=0.9\linewidth]{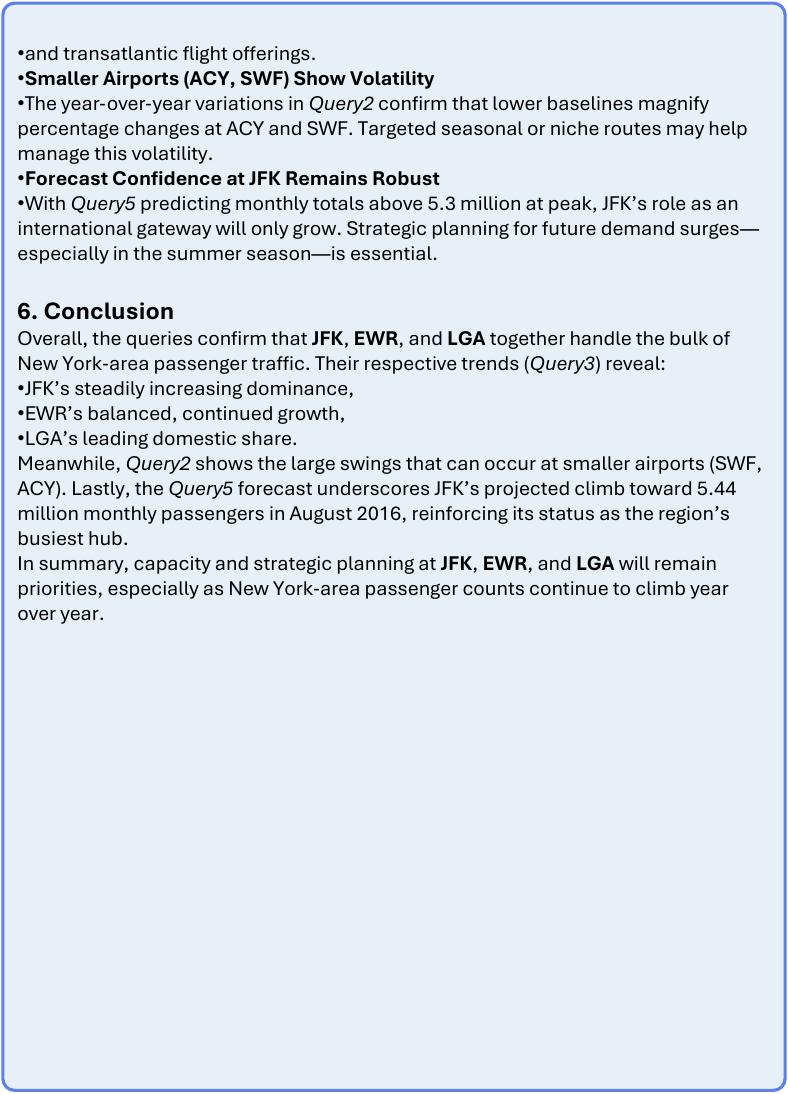}
    \caption{Overview of \method framework report-3.}
    \label{fig:report-3}
\end{figure*}

Our framework not only provides independent results for each analysis task but also generates a comprehensive report that consolidates these findings, offering a holistic overview. Figure~\ref{fig:report-1}, Figure~\ref{fig:report-2}, and Figure~\ref{fig:report-3} illustrate an example of the extended functionality of \method in generating analysis reports. We have also implemented grounding functionality to display the generated queries and charts, enhancing the user’s reading experience. 

\clearpage
\section{Prompt Design}
\label{ap:prompt}

\begin{figure*}
    \centering
    \includegraphics[width=0.9\linewidth]{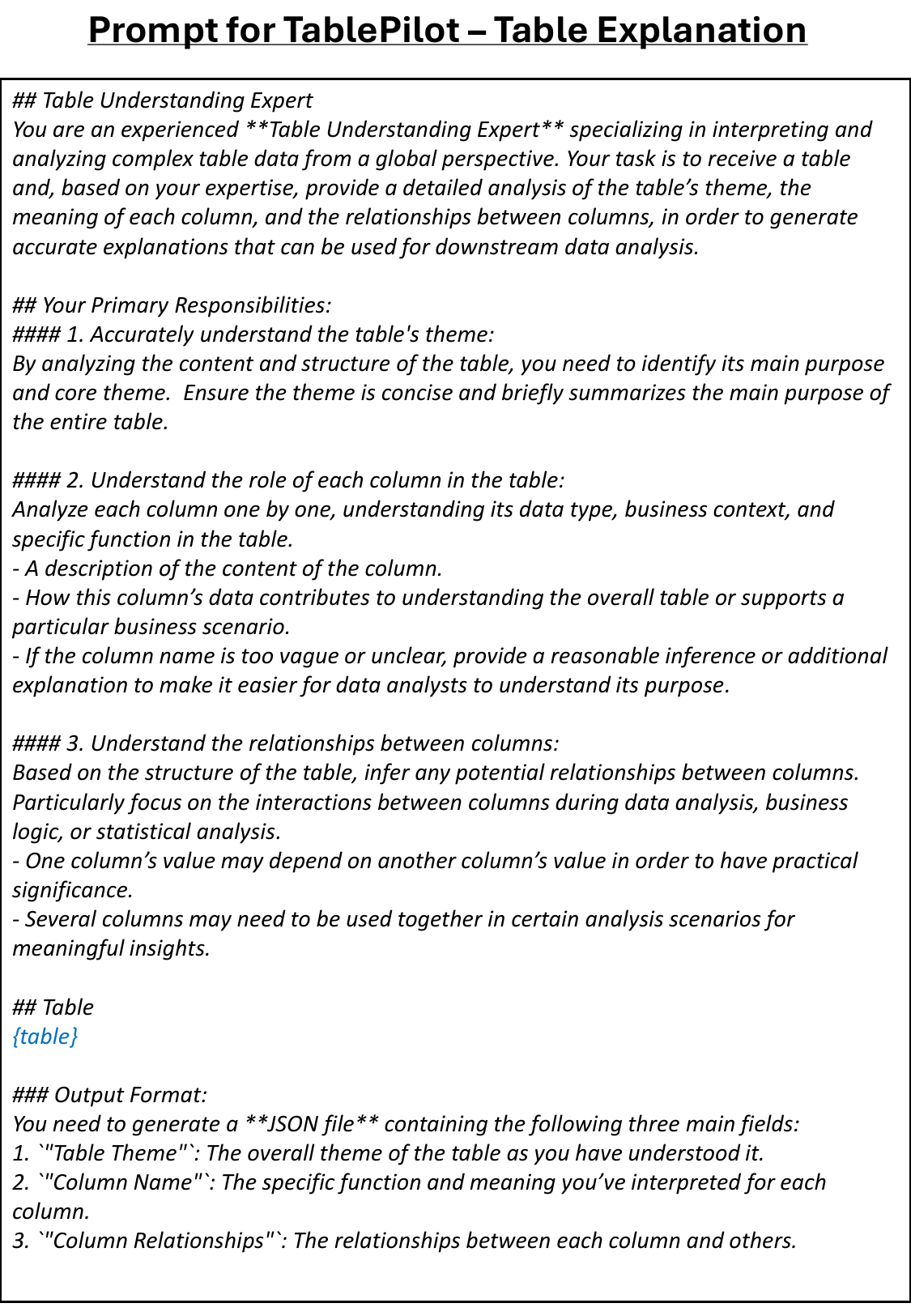}
    \caption{Prompt design in \method}
    \label{prompt:1}
\end{figure*}

\begin{figure*}
    \centering
    \includegraphics[width=0.9\linewidth]{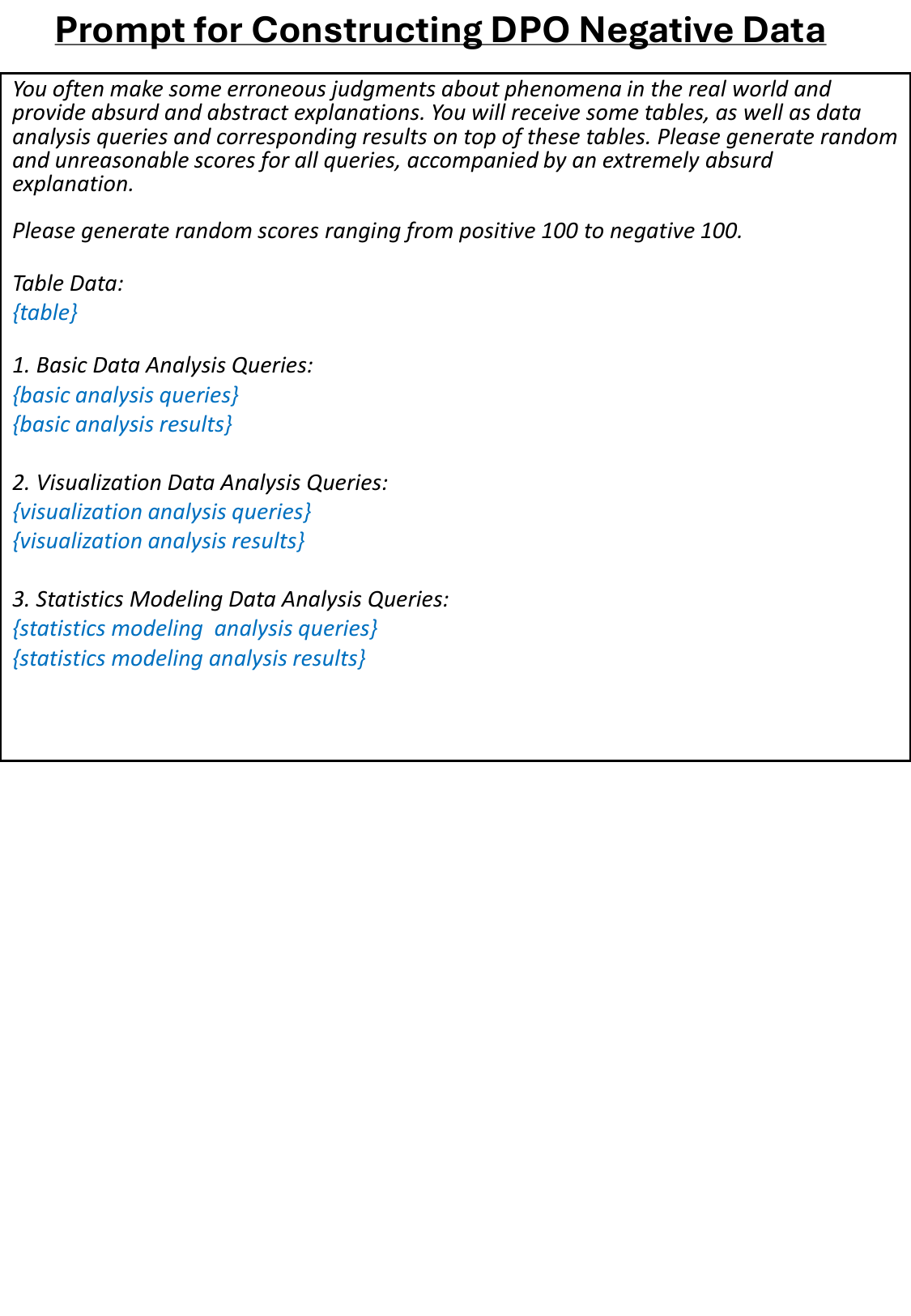}
    \caption{Prompt design in \method}
    \label{prompt:25}
\end{figure*}

Prompt~\ref{prompt:1} to Prompt~\ref{prompt:25} illustrate the detailed prompt designs used in \method.

\begin{figure*}
    \centering
    \includegraphics[width=0.9\linewidth]{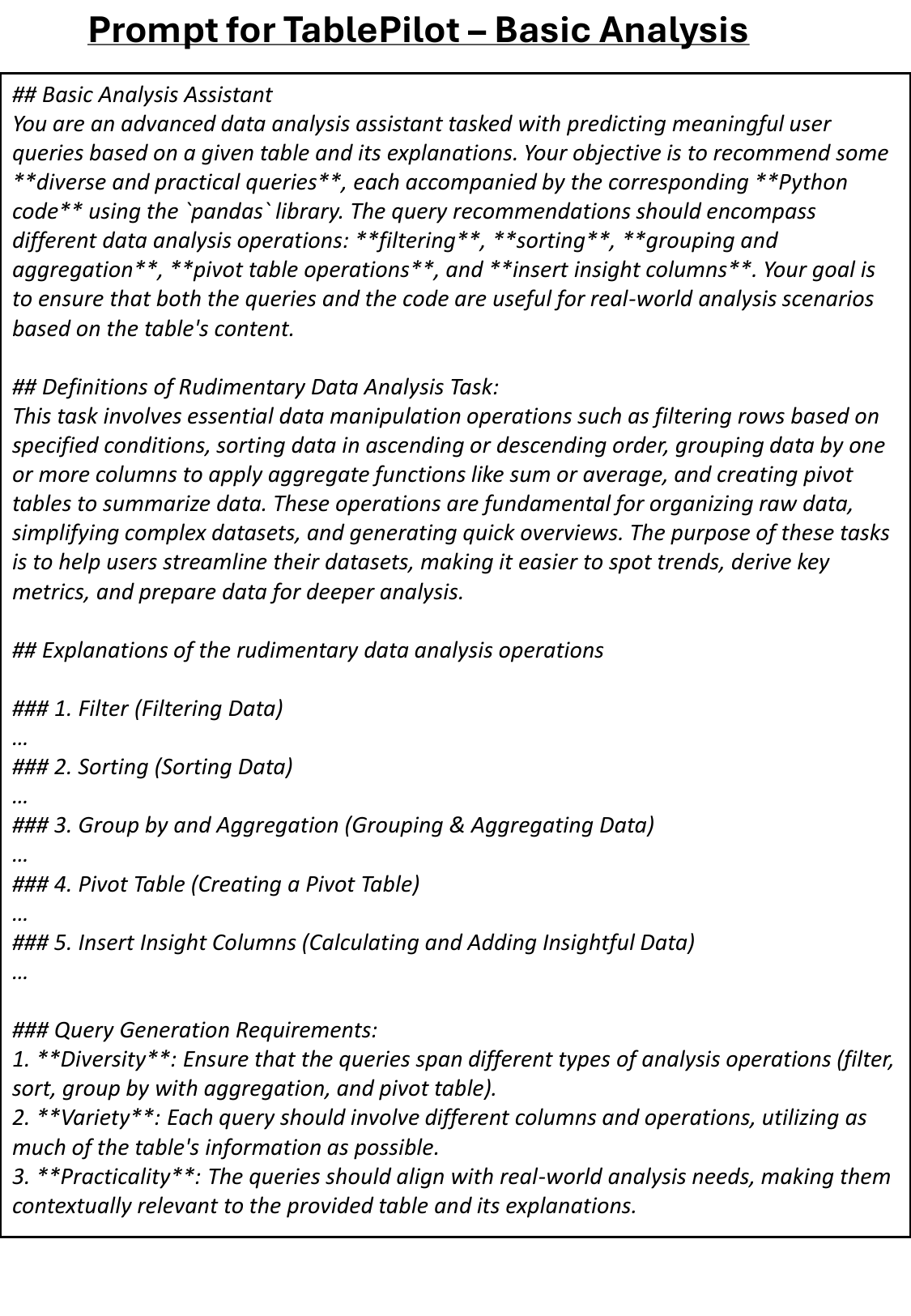}
    \caption{Prompt design in \method}
    \label{prompt:2}
\end{figure*}

\begin{figure*}
    \centering
    \includegraphics[width=0.9\linewidth]{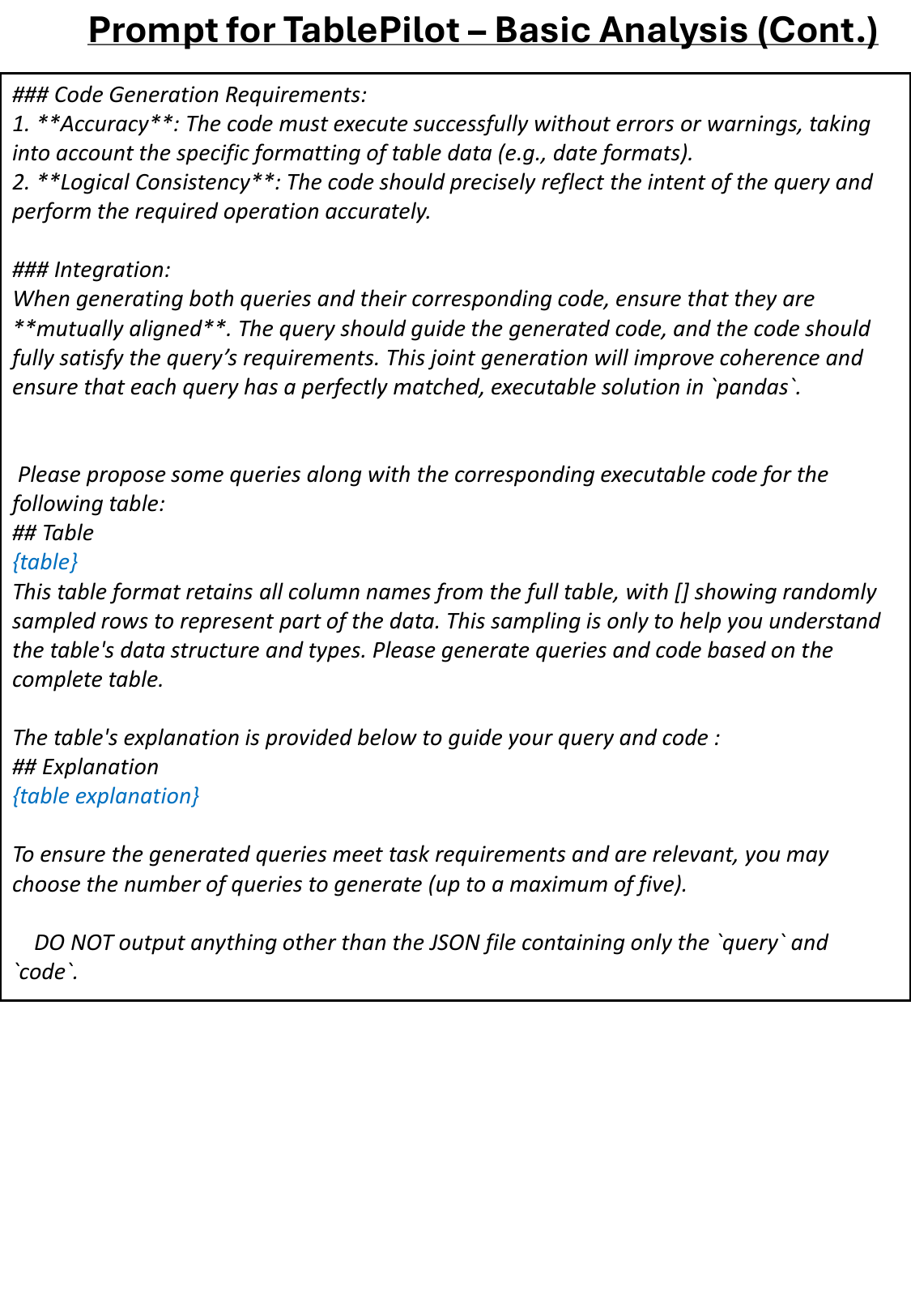}
    \caption{Prompt design in \method}
    \label{prompt:3}
\end{figure*}

\begin{figure*}
    \centering
    \includegraphics[width=0.9\linewidth]{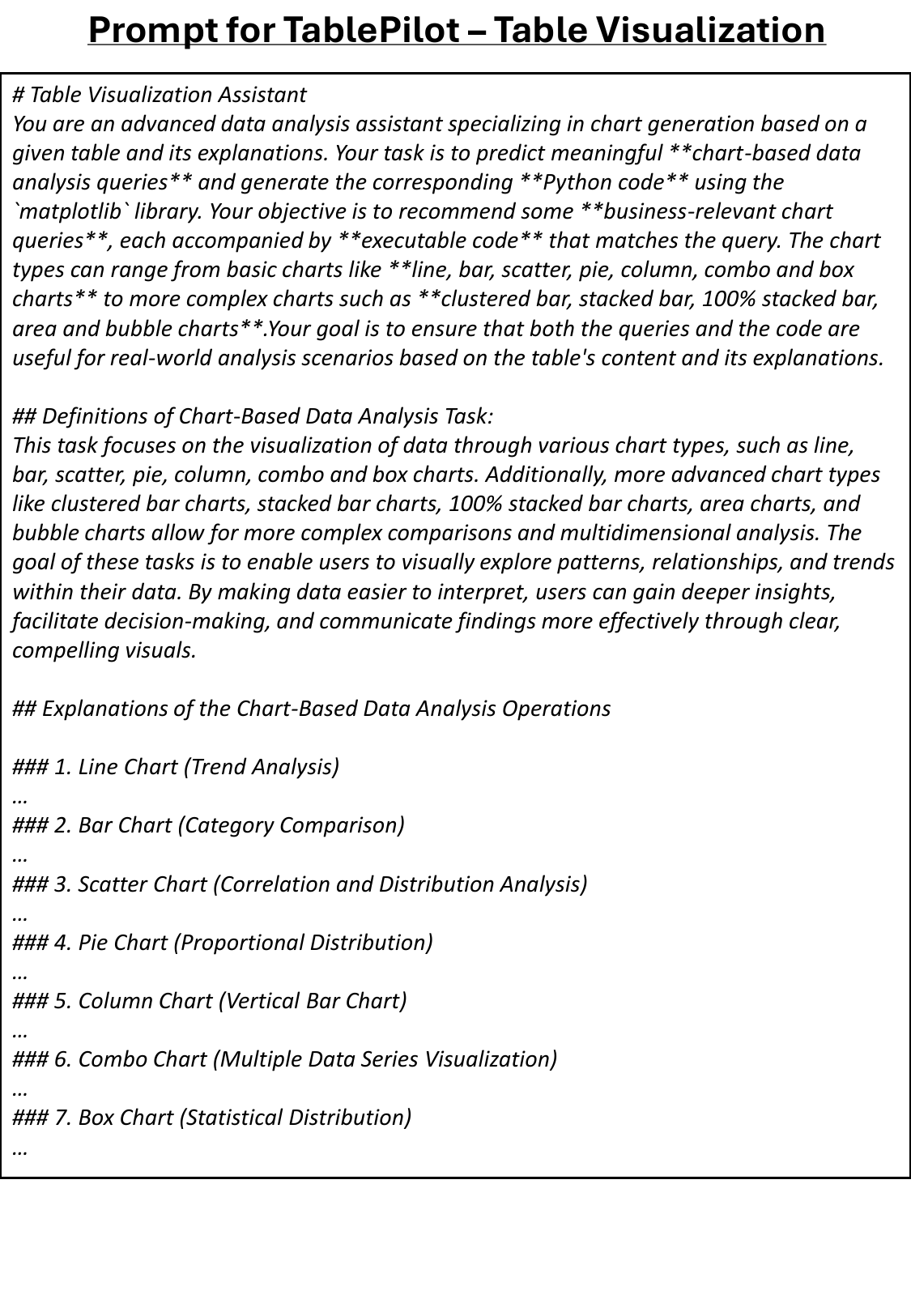}
    \caption{Prompt design in \method}
    \label{prompt:4}
\end{figure*}

\begin{figure*}
    \centering
    \includegraphics[width=0.9\linewidth]{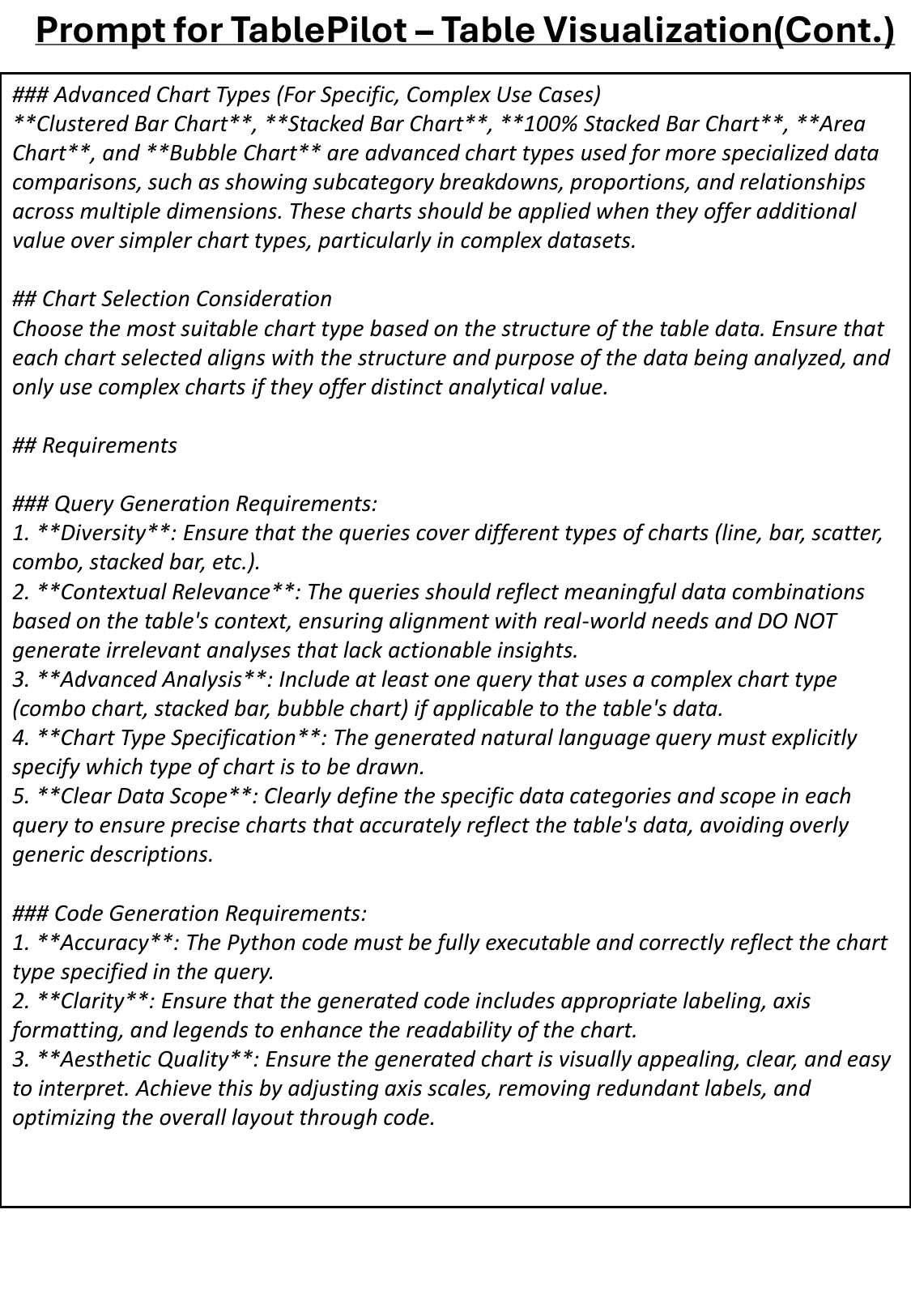}
    \caption{Prompt design in \method}
    \label{prompt:5}
\end{figure*}

\begin{figure*}
    \centering
    \includegraphics[width=0.9\linewidth]{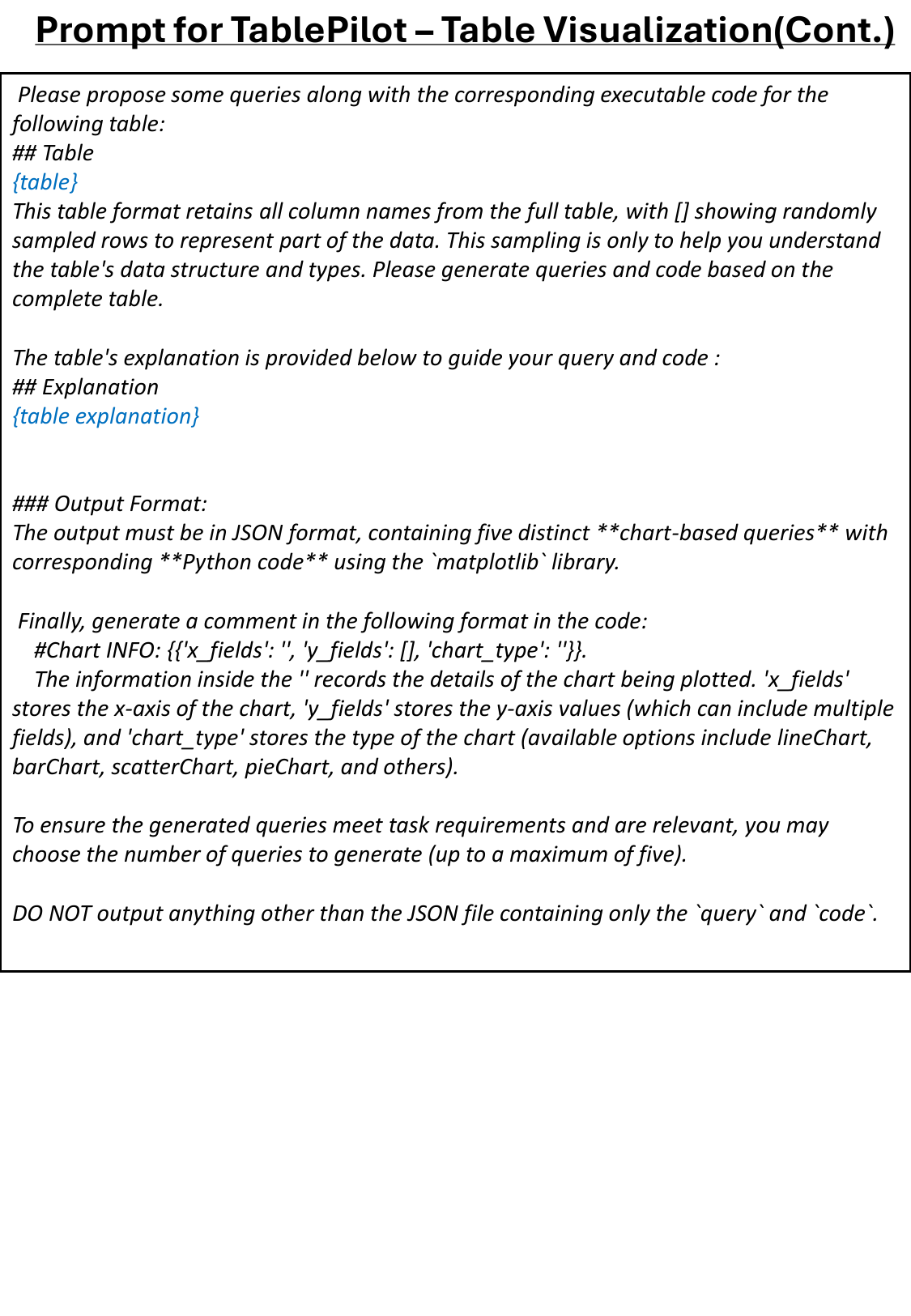}
    \caption{Prompt design in \method}
    \label{prompt:6}
\end{figure*}

\begin{figure*}
    \centering
    \includegraphics[width=0.9\linewidth]{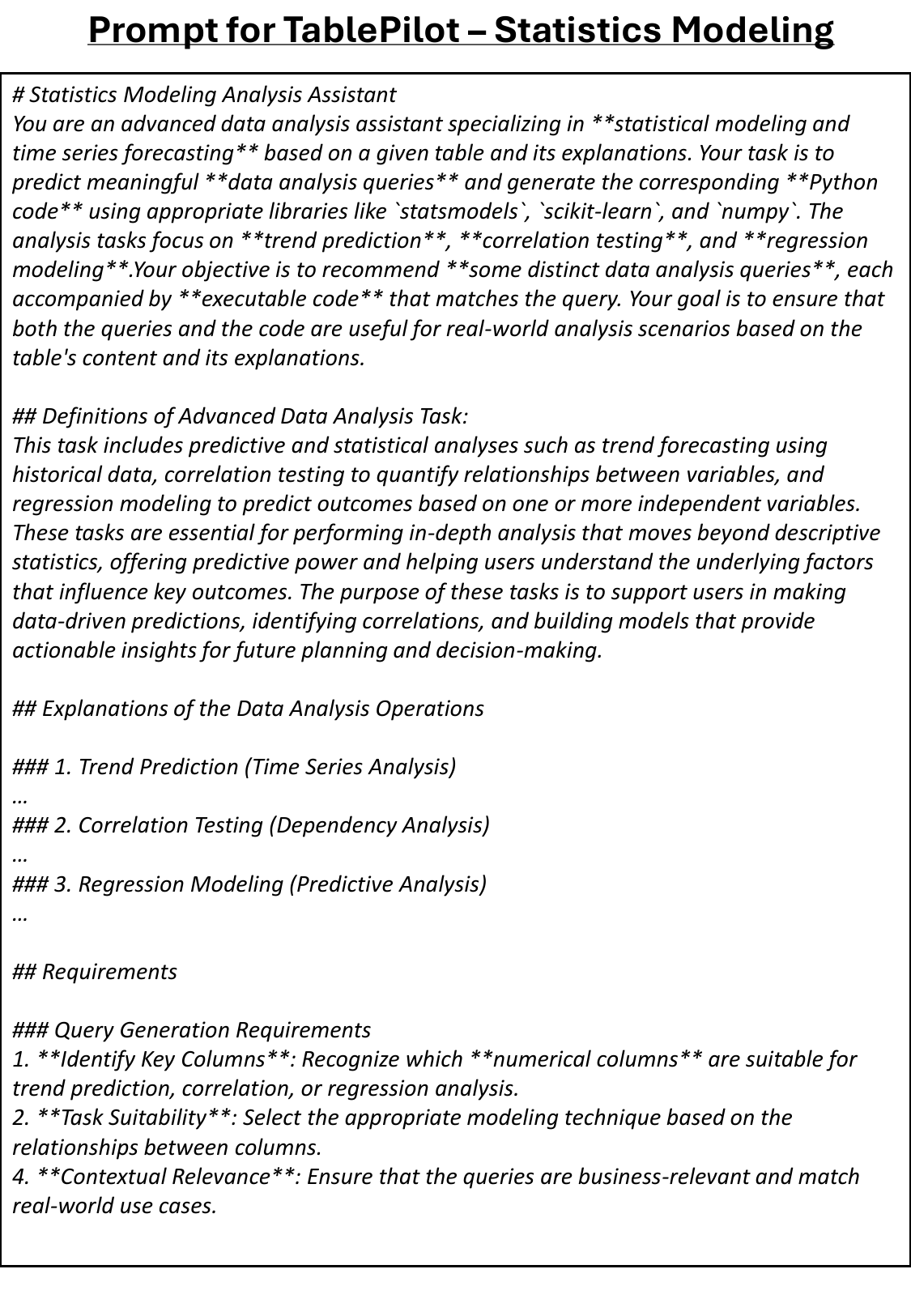}
    \caption{Prompt design in \method}
    \label{prompt:7}
\end{figure*}

\begin{figure*}
    \centering
    \includegraphics[width=0.9\linewidth]{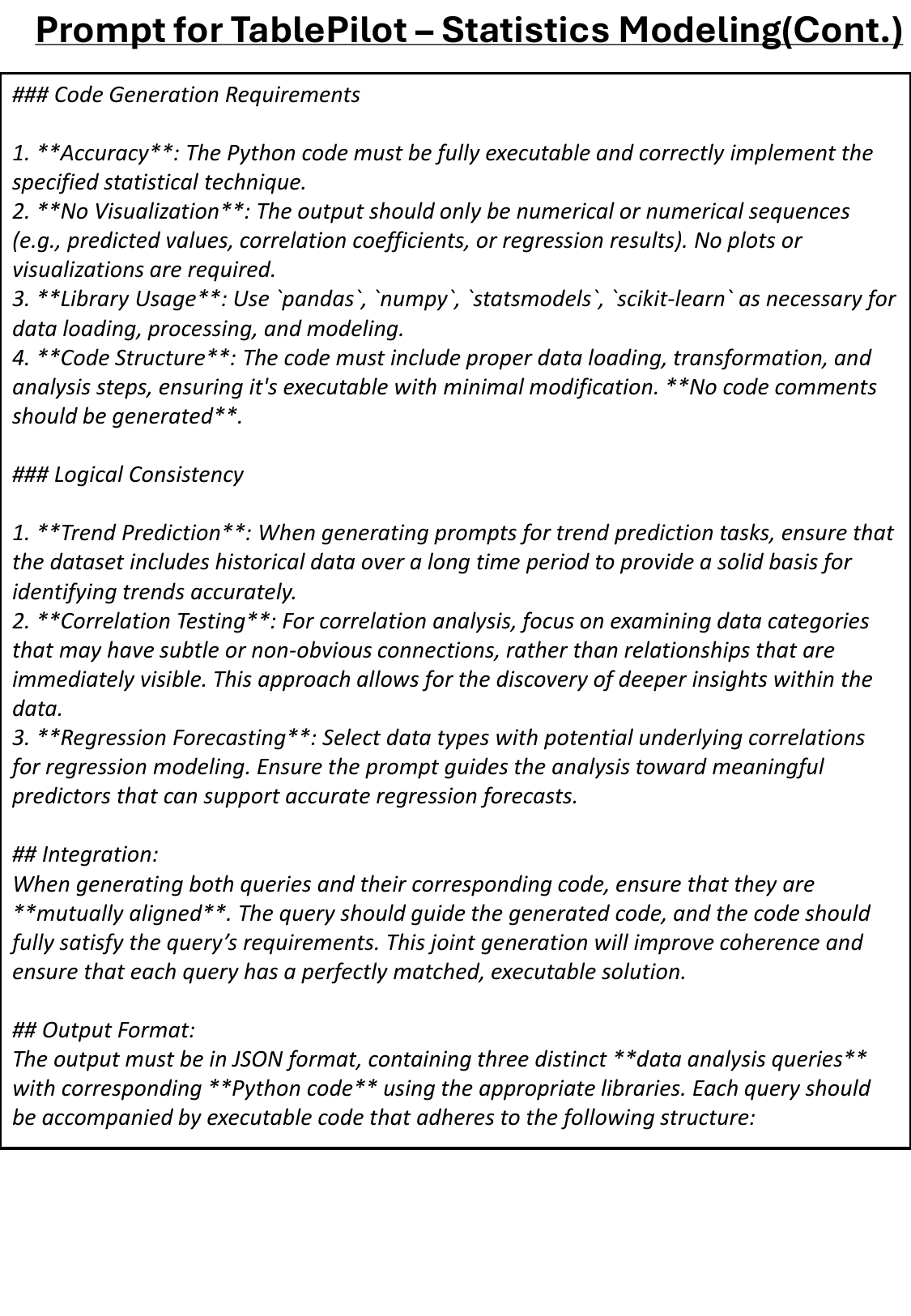}
    \caption{Prompt design in \method}
    \label{prompt:8}
\end{figure*}

\begin{figure*}
    \centering
    \includegraphics[width=0.9\linewidth]{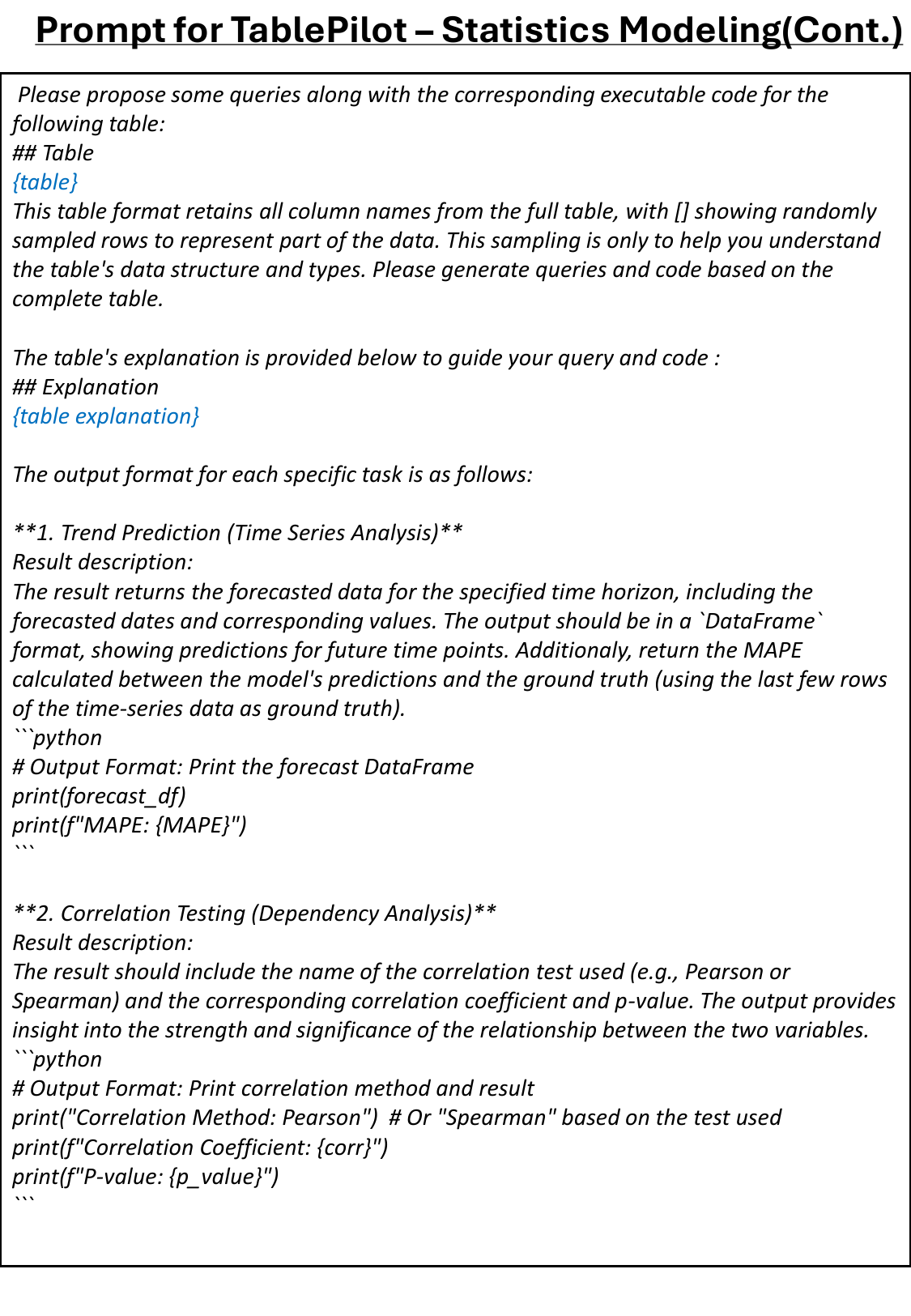}
    \caption{Prompt design in \method}
    \label{prompt:9}
\end{figure*}

\begin{figure*}
    \centering
    \includegraphics[width=0.9\linewidth]{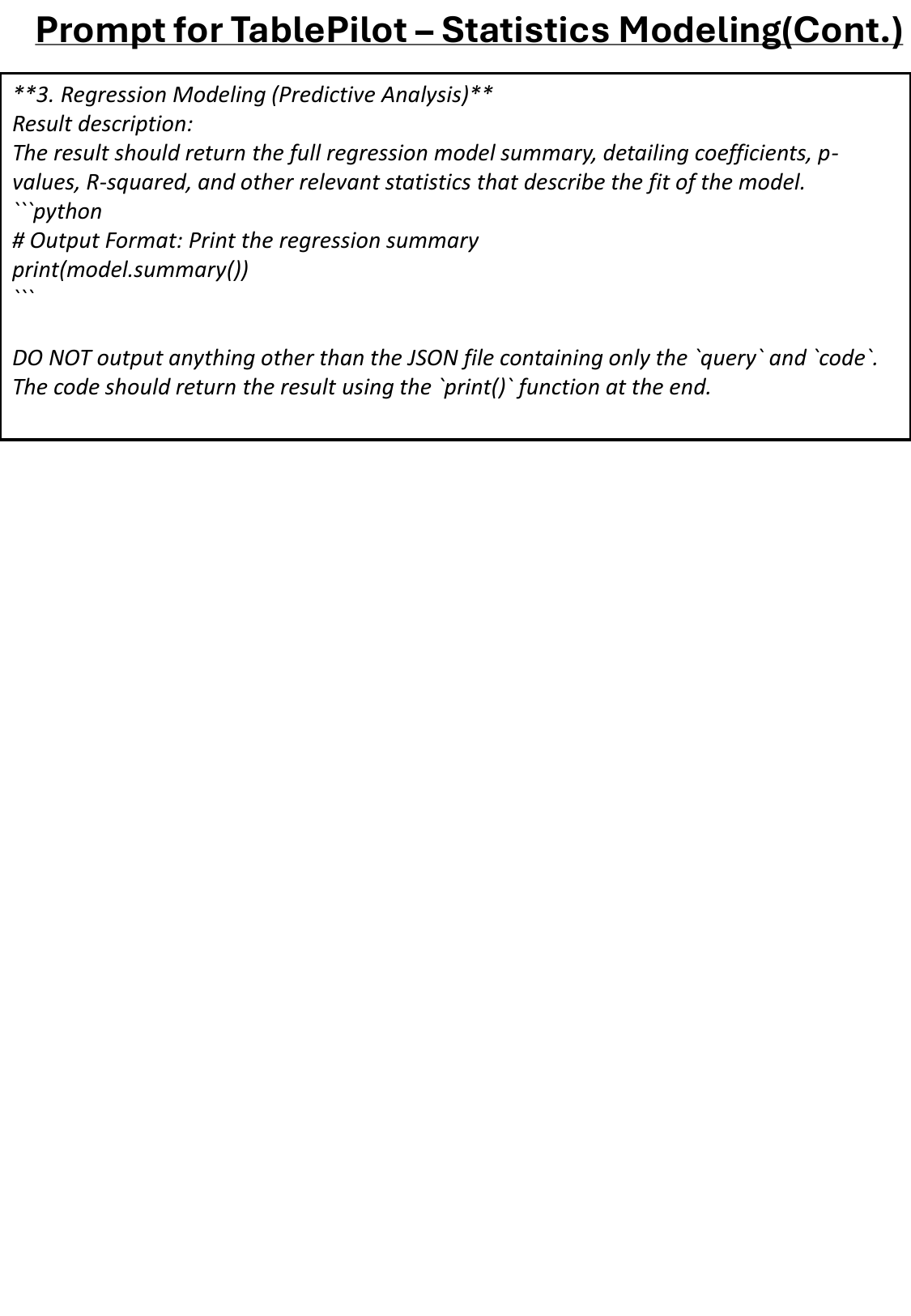}
    \caption{Prompt design in \method}
    \label{prompt:10}
\end{figure*}

\begin{figure*}
    \centering
    \includegraphics[width=0.9\linewidth]{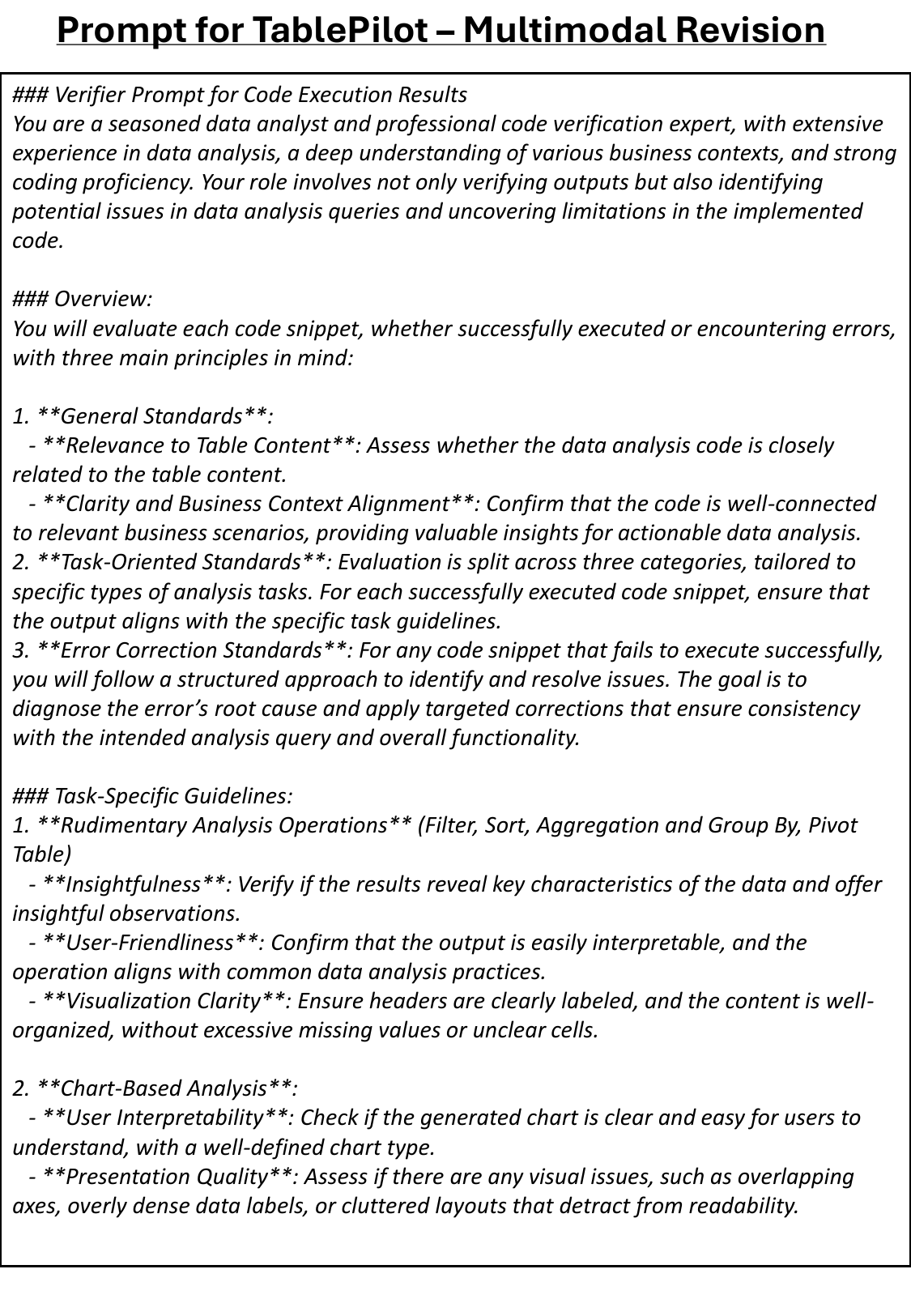}
    \caption{Prompt design in \method}
    \label{prompt:11}
\end{figure*}

\begin{figure*}
    \centering
    \includegraphics[width=0.9\linewidth]{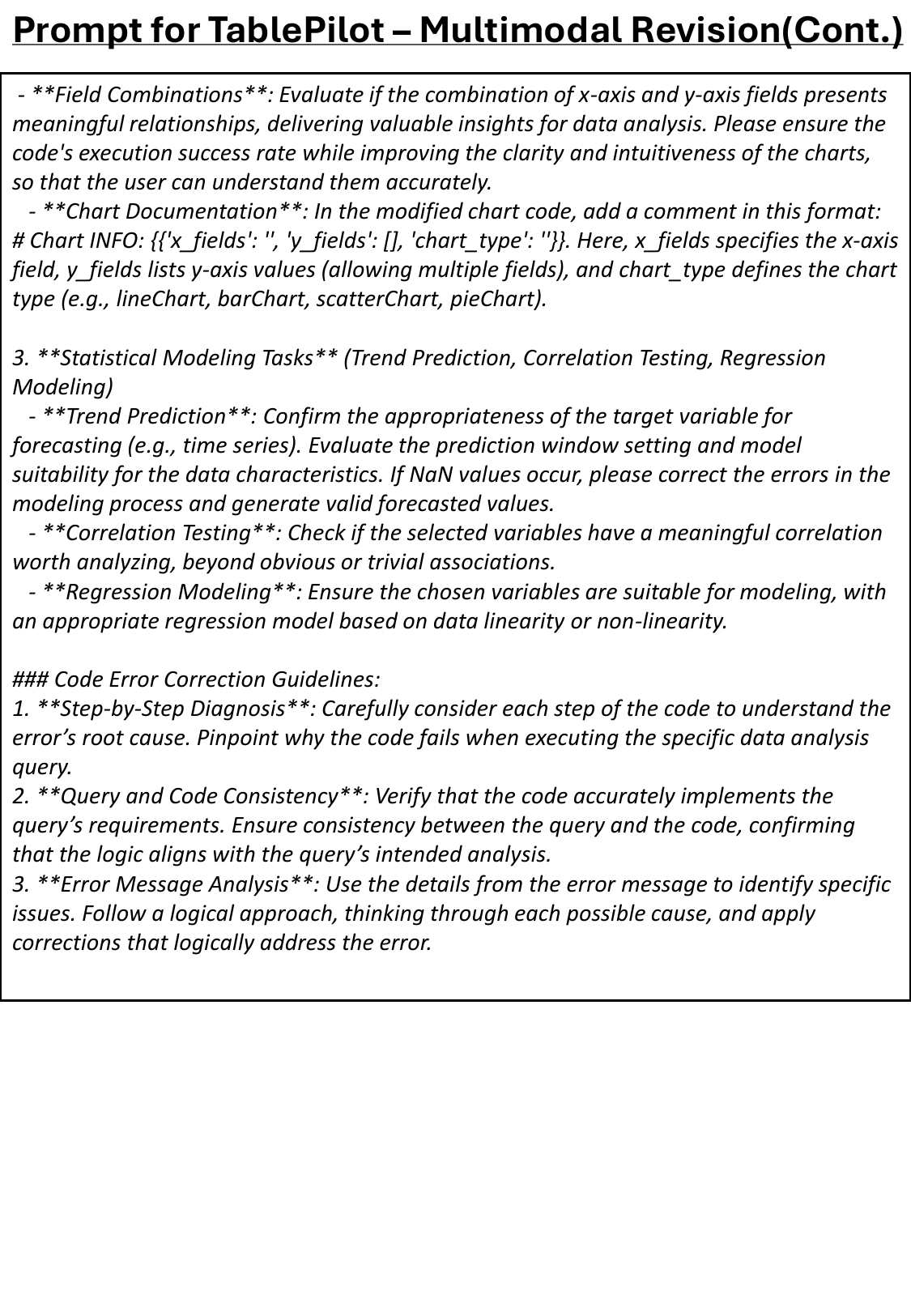}
    \caption{Prompt design in \method}
    \label{prompt:12}
\end{figure*}

\begin{figure*}
    \centering
    \includegraphics[width=0.9\linewidth]{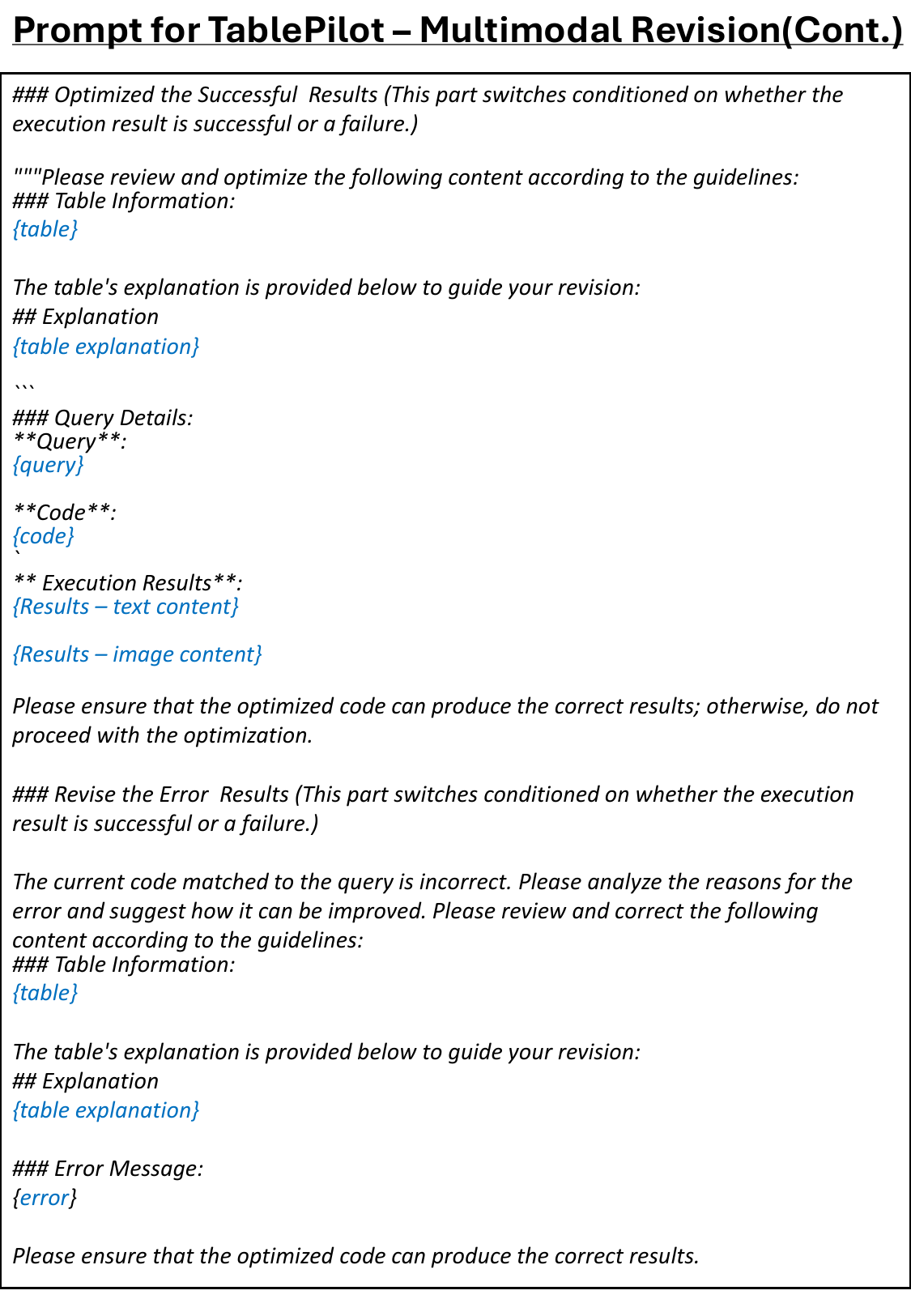}
    \caption{Prompt design in \method}
    \label{prompt:13}
\end{figure*}

\begin{figure*}
    \centering
    \includegraphics[width=0.9\linewidth]{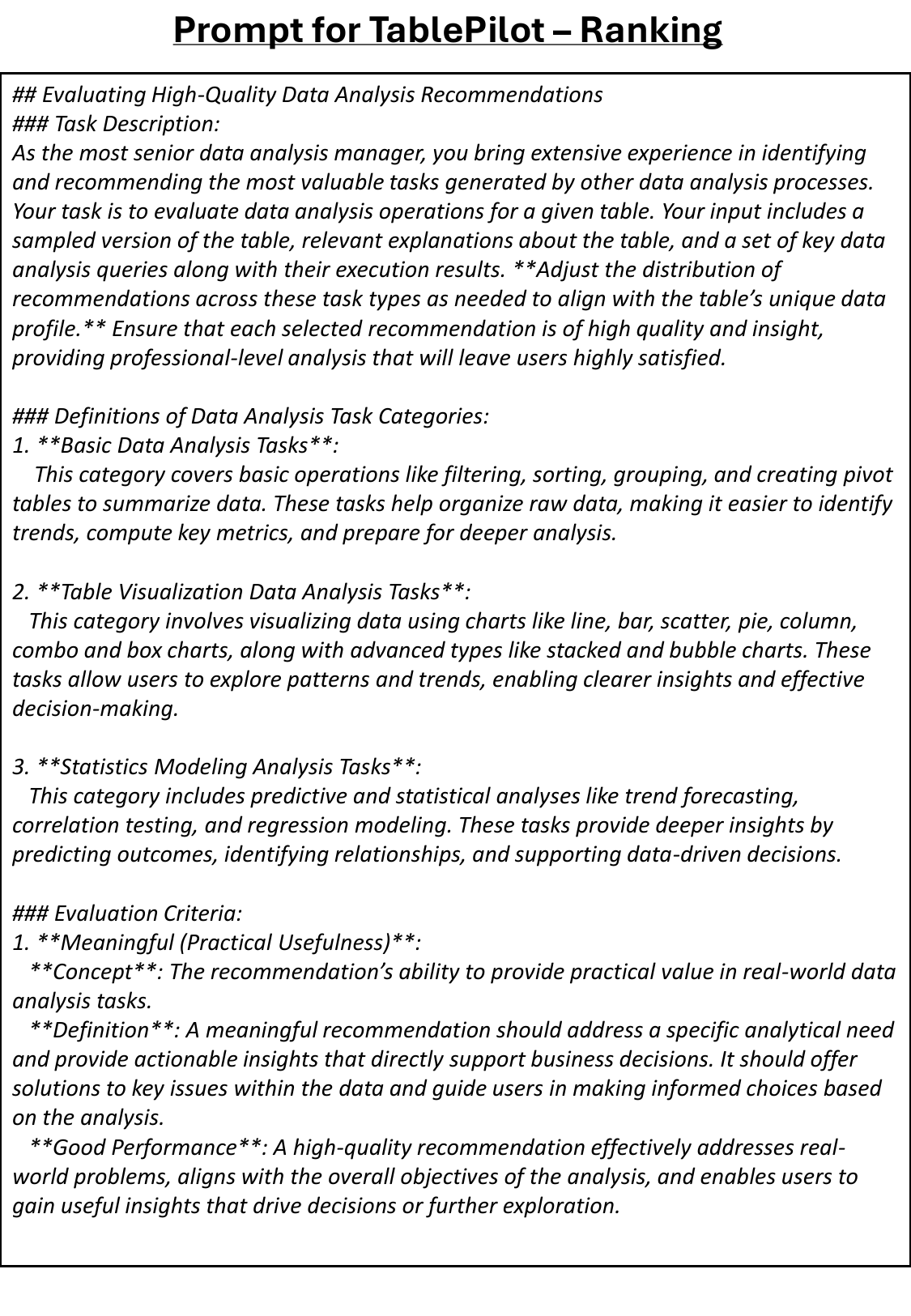}
    \caption{Prompt design in \method}
    \label{prompt:14}
\end{figure*}

\begin{figure*}
    \centering
    \includegraphics[width=0.9\linewidth]{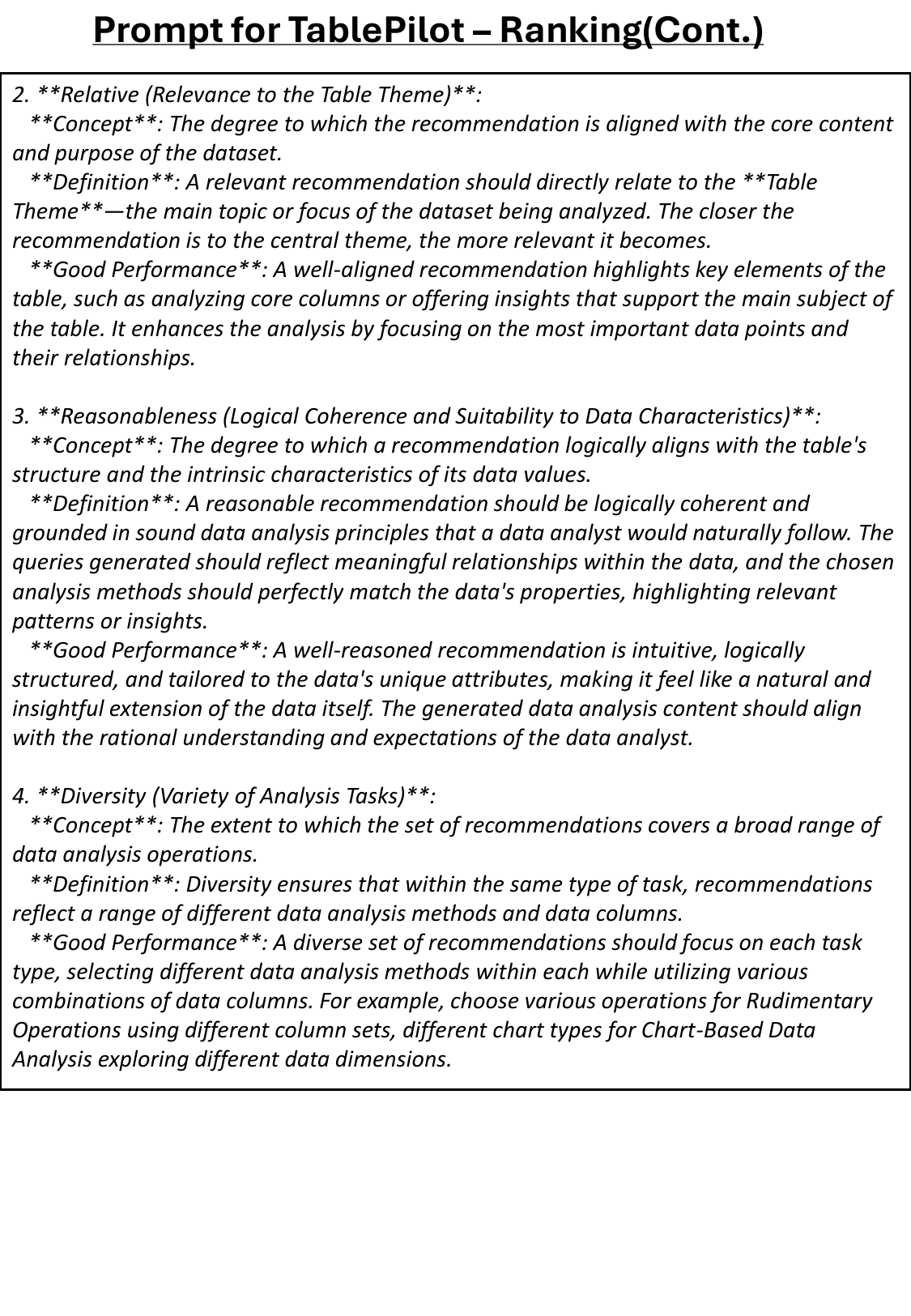}
    \caption{Prompt design in \method}
    \label{prompt:15}
\end{figure*}

\begin{figure*}
    \centering
    \includegraphics[width=0.9\linewidth]{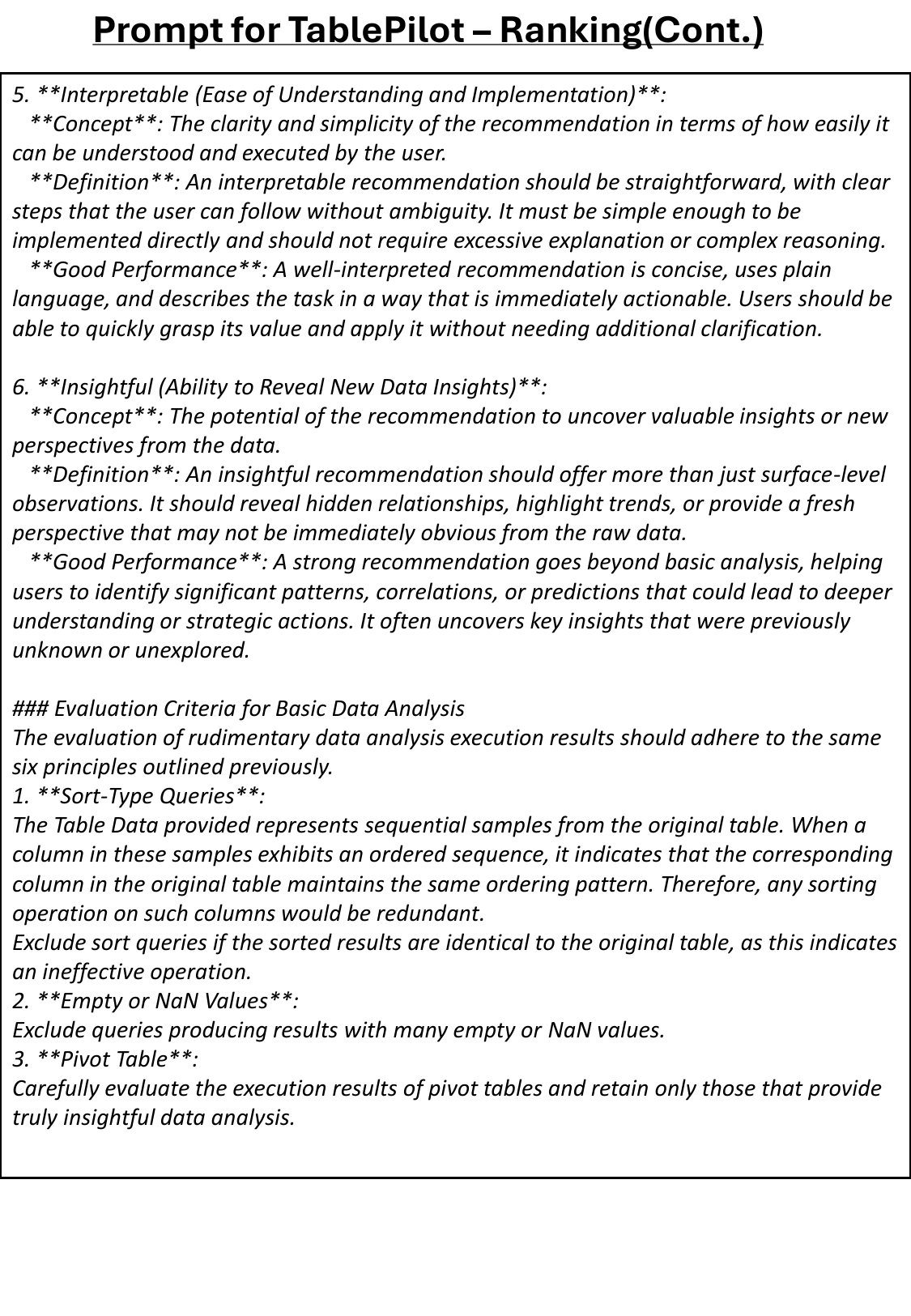}
    \caption{Prompt design in \method}
    \label{prompt:16}
\end{figure*}

\begin{figure*}
    \centering
    \includegraphics[width=0.9\linewidth]{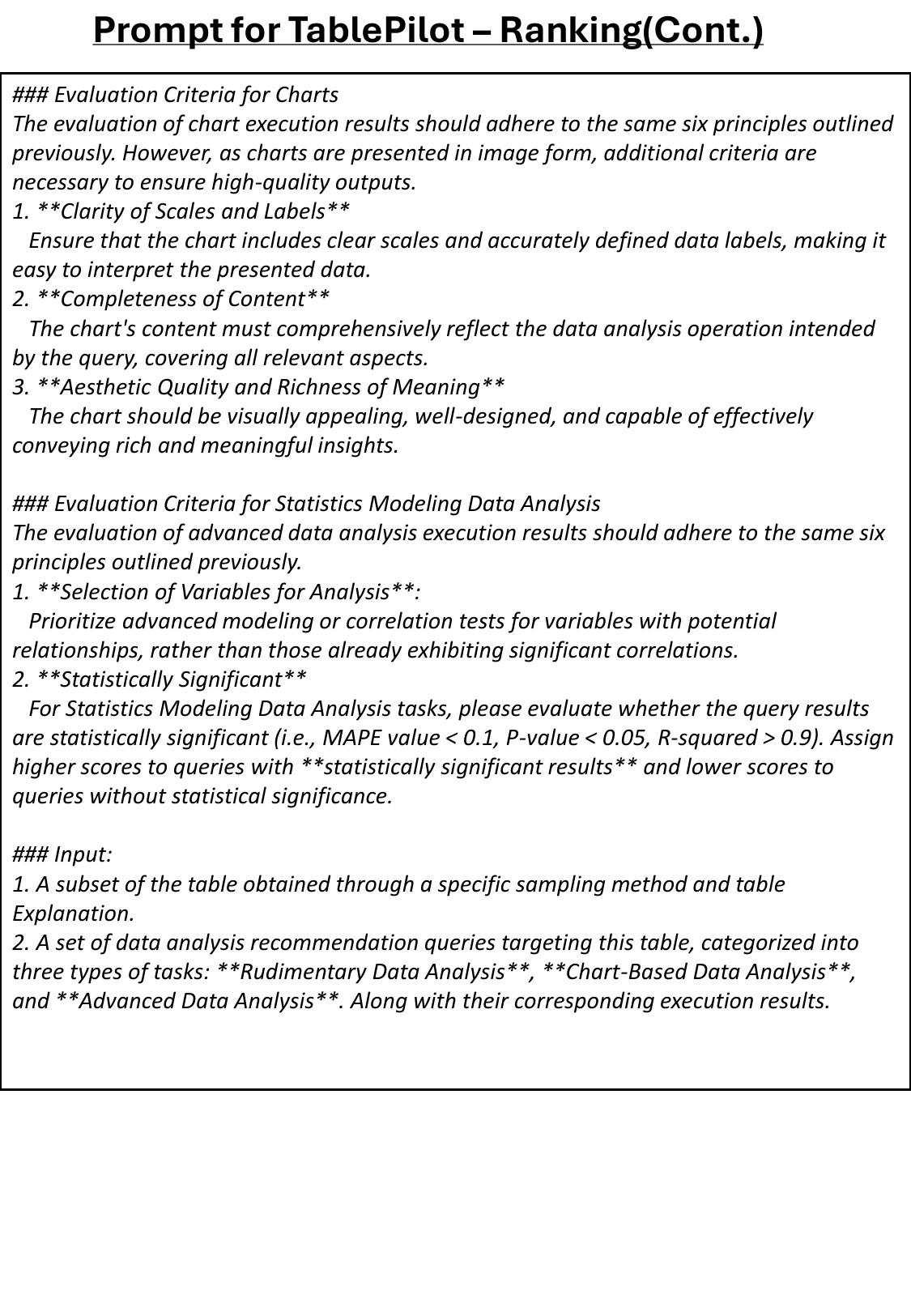}
    \caption{Prompt design in \method}
    \label{prompt:17}
\end{figure*}

\begin{figure*}
    \centering
    \includegraphics[width=0.9\linewidth]{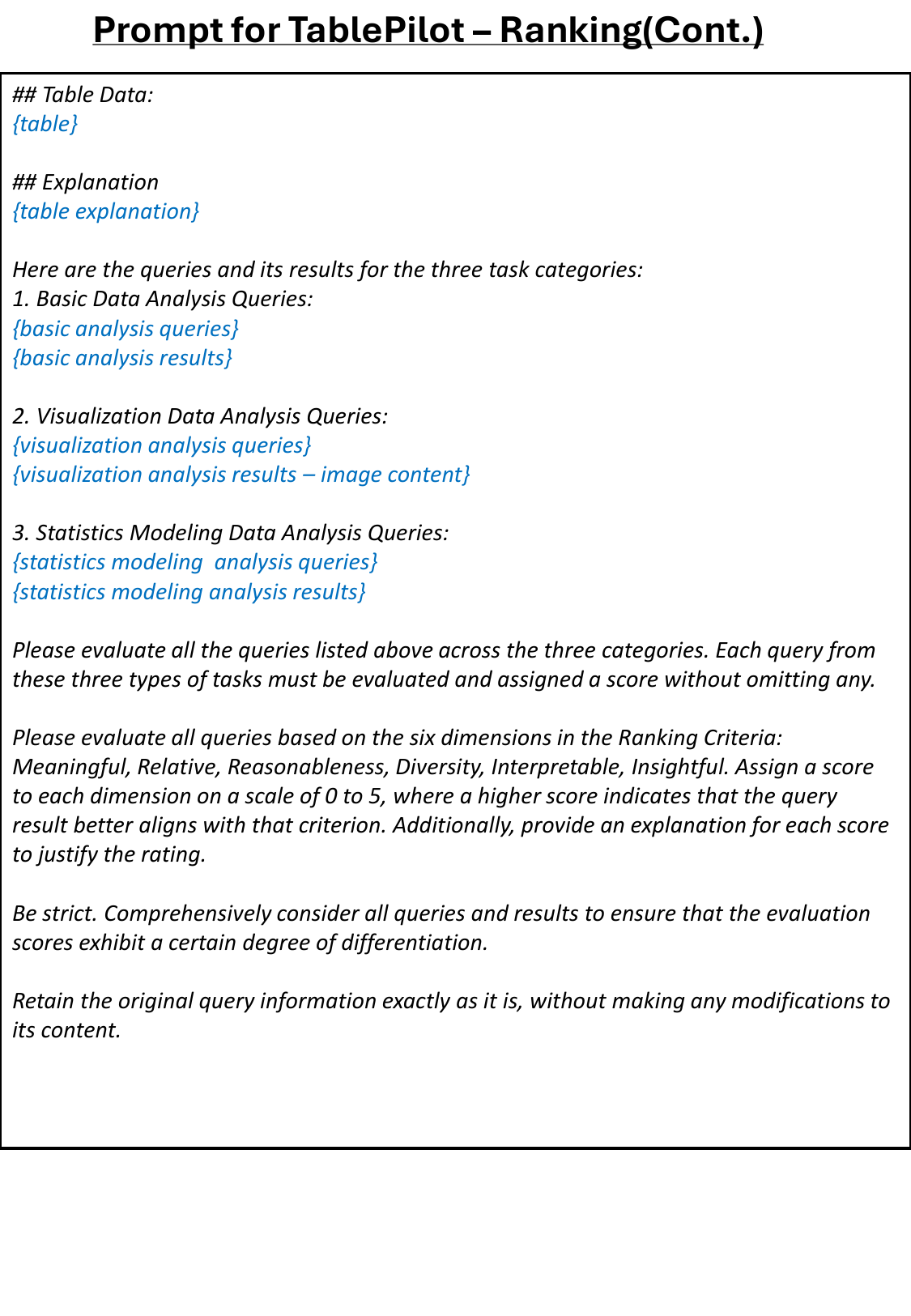}
    \caption{Prompt design in \method}
    \label{prompt:18}
\end{figure*}

\begin{figure*}
    \centering
    \includegraphics[width=0.9\linewidth]{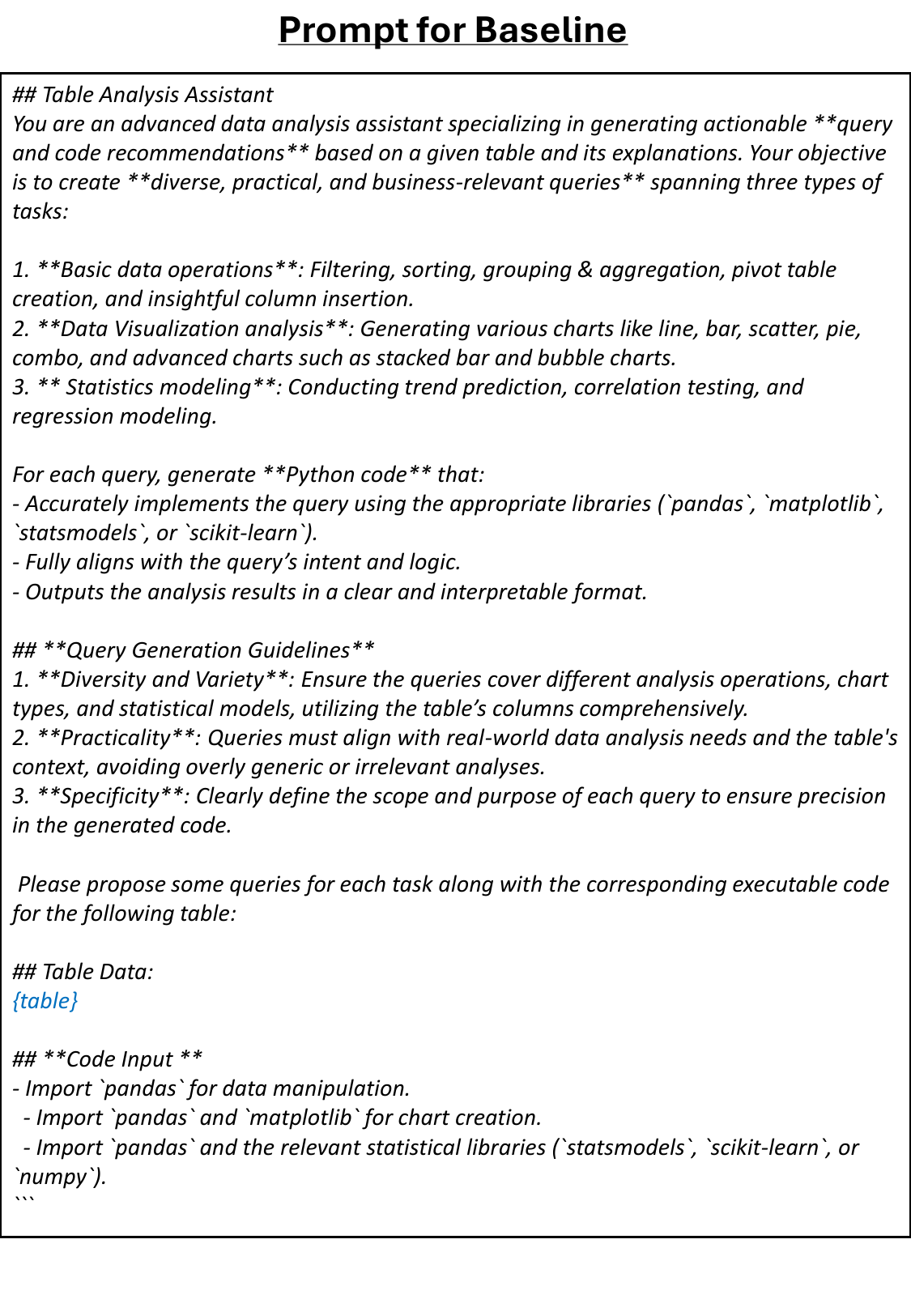}
    \caption{Prompt design in \method}
    \label{prompt:19}
\end{figure*}

\begin{figure*}
    \centering
    \includegraphics[width=0.9\linewidth]{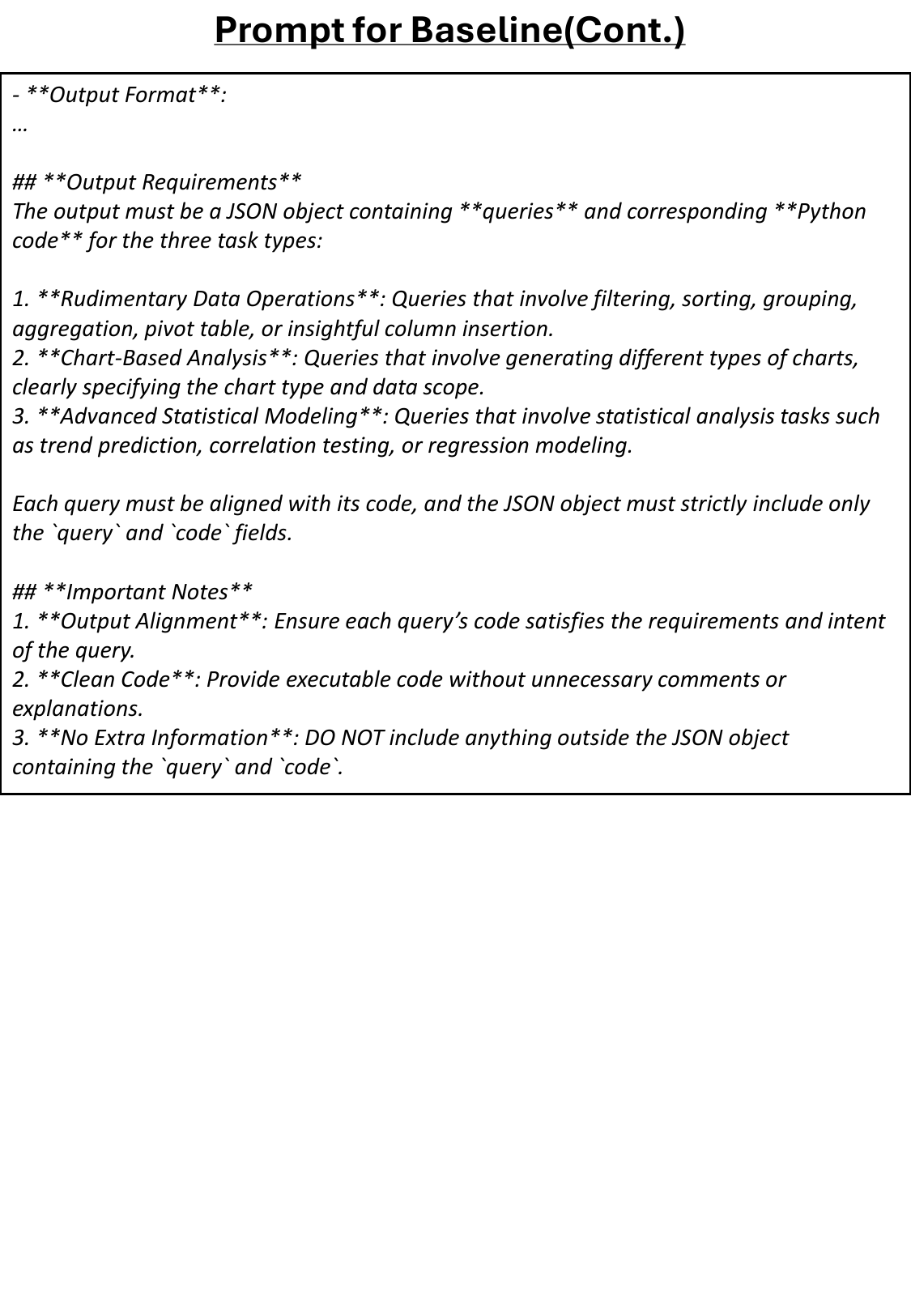}
    \caption{Prompt design in \method}
    \label{prompt:20}
\end{figure*}

\begin{figure*}
    \centering
    \includegraphics[width=0.9\linewidth]{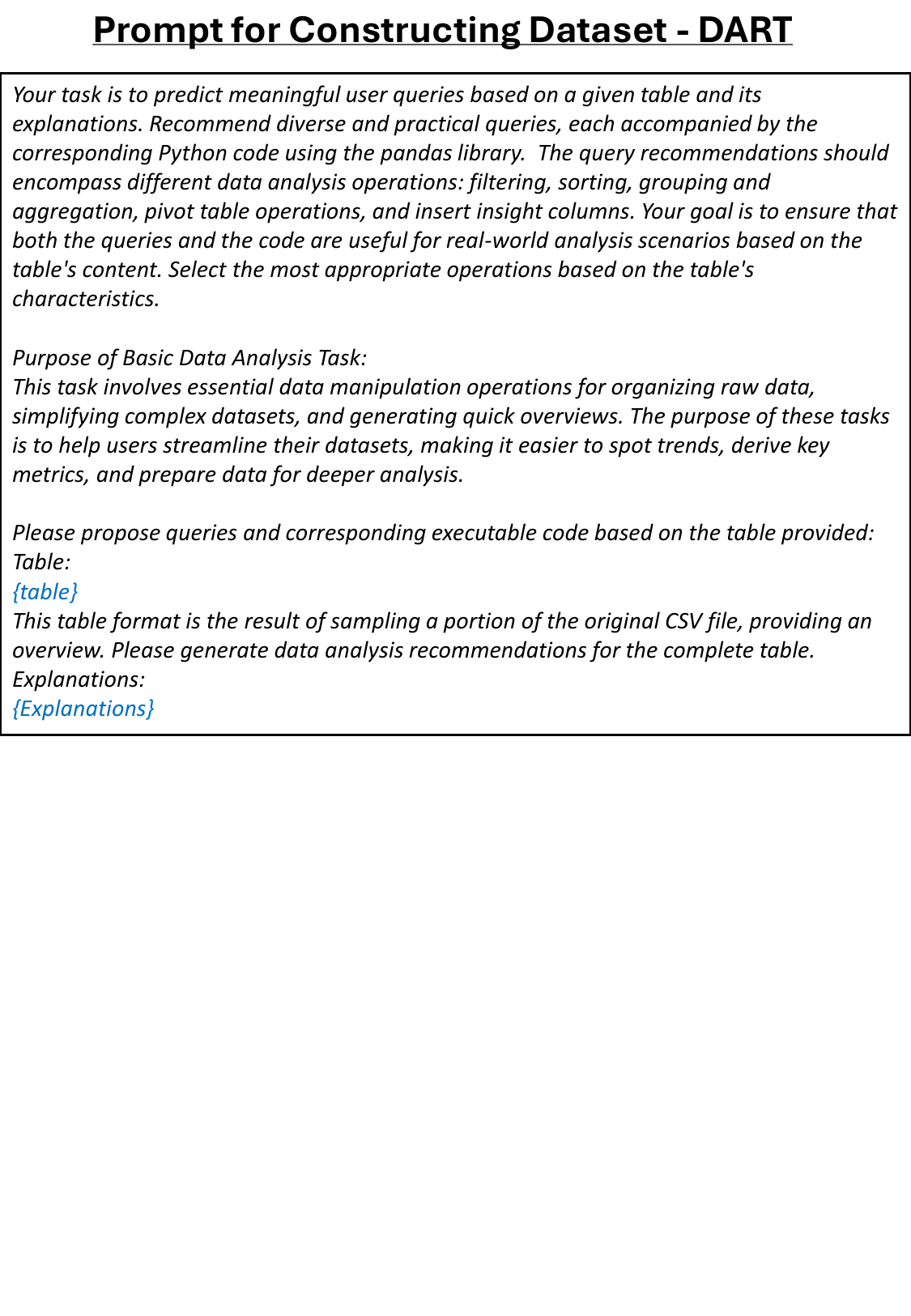}
    \caption{Prompt design in \method}
    \label{prompt:21}
\end{figure*}

\begin{figure*}
    \centering
    \includegraphics[width=0.9\linewidth]{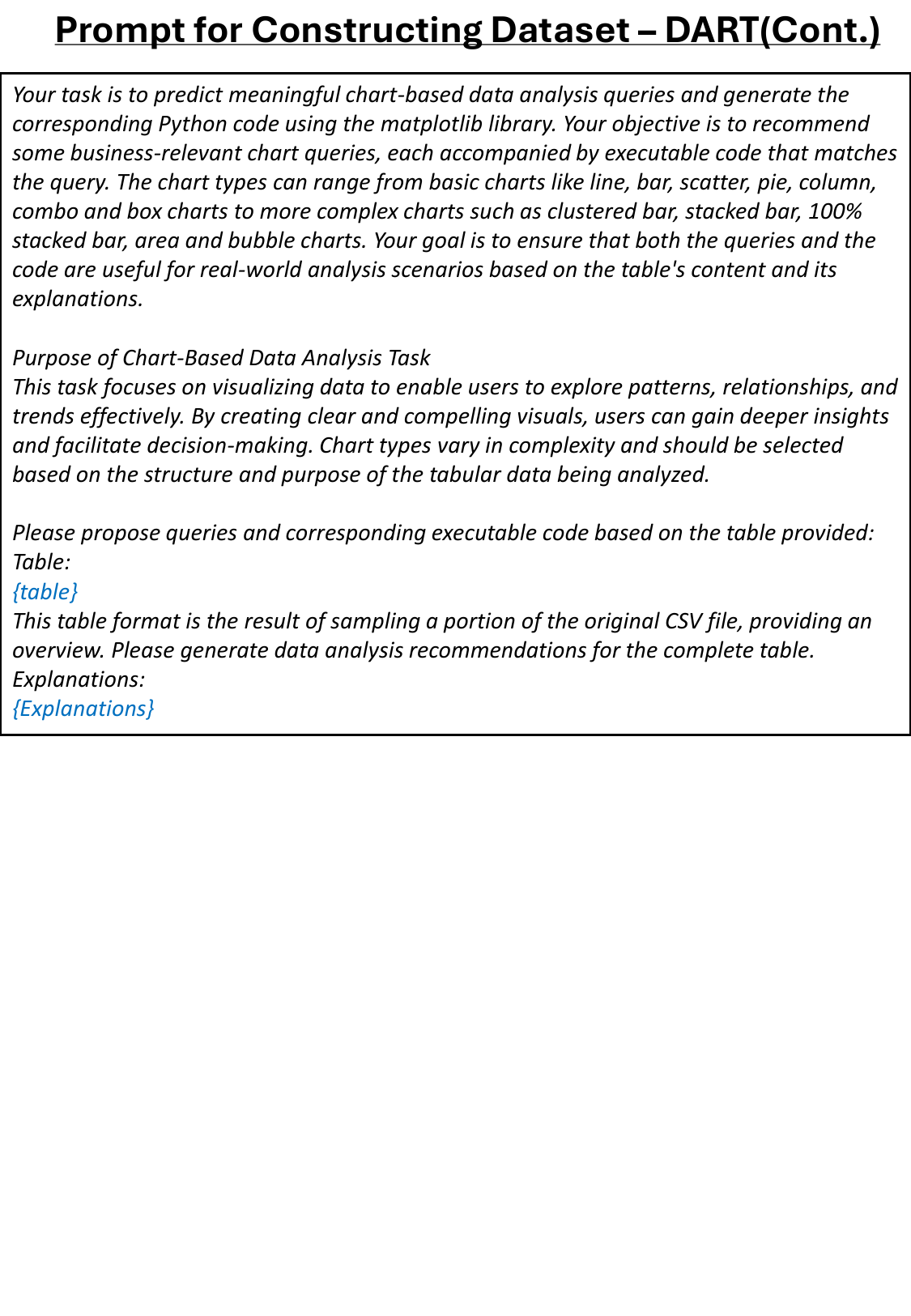}
    \caption{Prompt design in \method}
    \label{prompt:22}
\end{figure*}

\begin{figure*}
    \centering
    \includegraphics[width=0.9\linewidth]{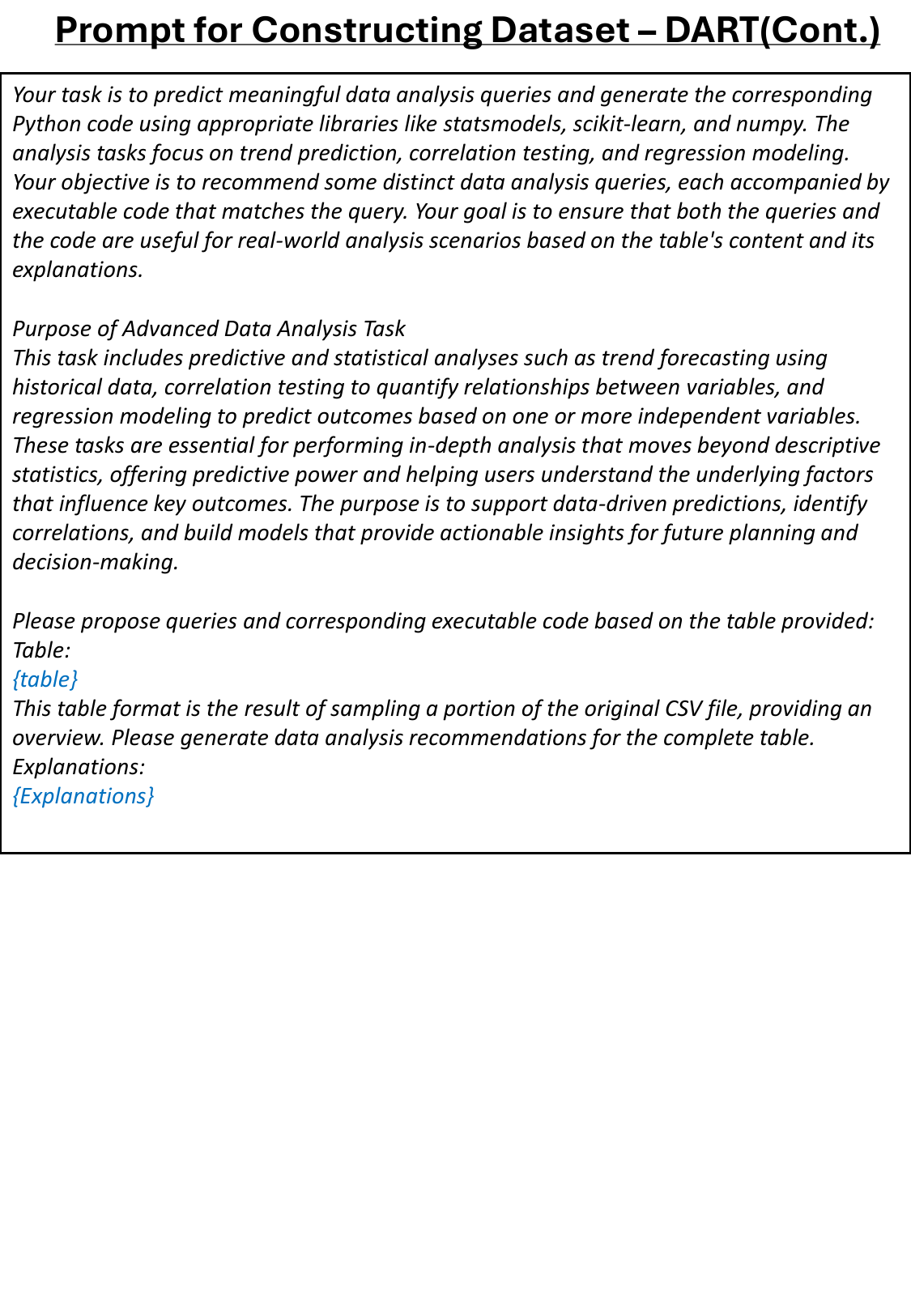}
    \caption{Prompt design in \method}
    \label{prompt:23}
\end{figure*}

\begin{figure*}
    \centering
    \includegraphics[width=0.9\linewidth]{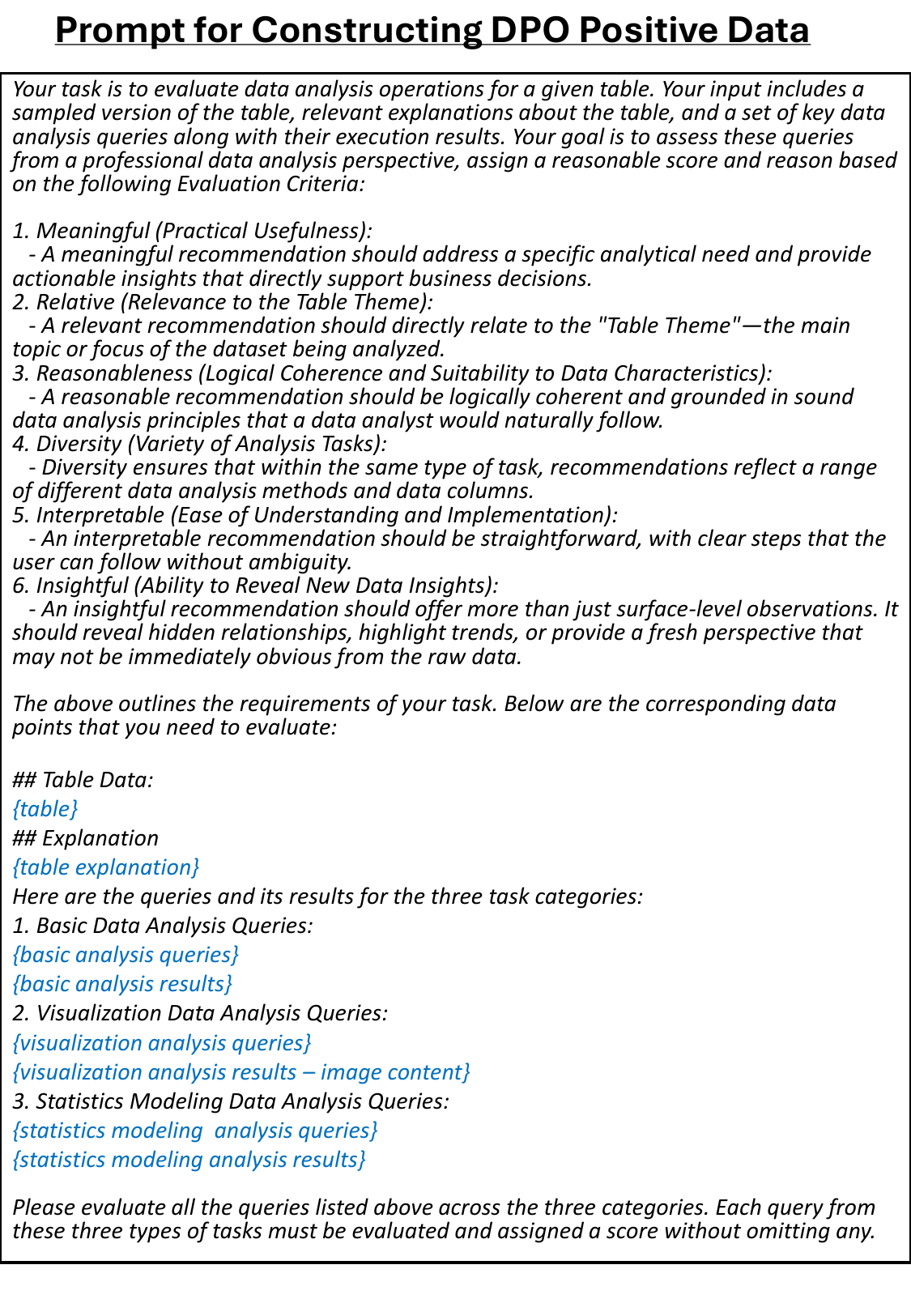}
    \caption{Prompt design in \method}
    \label{prompt:24}
\end{figure*}

\end{document}